\documentclass[%
notitlepage,
superscriptaddress,
longbibliography,
 amsmath,amssymb,
 aps,
]{revtex4-2}

\usepackage{amsmath}
\usepackage{bm}
\usepackage{graphicx}
\usepackage{dcolumn}
\usepackage{bm}
\usepackage{hyperref}
\usepackage{epstopdf}
\usepackage[mathlines]{lineno}
\usepackage{multirow}    
\usepackage{booktabs}    
\usepackage{float}
\usepackage{subcaption}
\usepackage{changes}
\definechangesauthor[name=Lucas, color=red]{LS}
\usepackage{multirow}
\usepackage{placeins}
\usepackage{url}
\makeatletter
\newcommand{\doi@}[1]{\href{https://doi.org/#1}{#1}}
\DeclareRobustCommand{\doi}{\hyper@normalise\doi@}
\makeatother
\begin{document}
\title{Learning Mesh-Free Discrete Differential Operators \\ with Self-Supervised Graph Neural Networks}

\author{Lucas Gerken Starepravo}
\email{lucas.gerkenstarepravo@postgrad.manchester.ac.uk}
\affiliation{School of Engineering, The University of Manchester, Manchester, UK}

\author{Georgios Fourtakas}
\affiliation{School of Engineering, The University of Manchester, Manchester, UK}

\author{Steven~Lind}
\affiliation{School of Engineering, Cardiff University, Cardiff, UK}

\author{Ajay B. Harish}
\affiliation{School of Engineering, The University of Manchester, Manchester, UK}

\author{Tianning Tang}
\affiliation{School of Engineering, The University of Manchester, Manchester, UK}

\author{Jack R. C. King}
\affiliation{School of Engineering, The University of Manchester, Manchester, UK}

\begin{abstract}
\vspace{0.5cm}\noindent
Mesh-free numerical methods provide flexible discretisations for complex geometries; however, classical meshless discrete differential operators typically trade low computational cost for limited accuracy or high accuracy for substantial per-stencil computation. We introduce a parametrised framework for learning mesh-free discrete differential operators using a graph neural network trained via polynomial moment constraints derived from truncated Taylor expansions. The model maps local stencils relative positions directly to discrete operator weights. The current work demonstrates that neural networks can learn classical polynomial consistency while retaining robustness to irregular neighbourhood geometry. The learned operators depend only on local geometry, are resolution-agnostic, and can be reused across particle configurations and governing equations. We evaluate the framework using standard numerical analysis diagnostics, showing improved accuracy over Smoothed Particle Hydrodynamics, and a favourable accuracy–cost trade-off relative to a representative high-order consistent mesh-free method in the moderate-accuracy regime. Applicability is demonstrated by solving the weakly compressible Navier–Stokes equations using the learned operators.

\end{abstract}

\maketitle

\section{Introduction}
Partial differential equations (PDEs) are foundational to the modelling of physical systems across science and engineering. Yet, analytical solutions are rarely available for realistic problems of interest. Thus, numerical methods are often used to approximate the governing equations. 

Mesh-based numerical methods, such as the finite difference method (FDM)~\cite{Lele1992}, finite element method (FEM)~\cite{Clough1990}, finite volume method (FVM)~\citep{mcdonald1971computation, maccormack1972computational}, and spectral methods~\cite{Gottlieb1977}, have long been the standard tools for solving PDEs~\cite{Vacondio2021}, given these methods can be made extremely accurate for simple geometries. Despite their maturity, these approaches face challenges regarding accuracy near interfaces~\cite{Lind2020} and the overhead associated with the mesh: this includes both the need for complex adaptation procedures~\citep{Liu2005} and initial generation phases that, for intricate geometries, may be more time-consuming than the simulation itself~\cite{wood_porous}. Mesh-free numerical methods have significant potential, as discretisation relies solely on local connectivity information between collocation points without requiring topological connectivity. Furthermore, collocations points can be positioned flexibly to satisfy resolution requirements and conform to complex geometries~\citep{Liu2005,Vacondio2021}. 

Developed in the 1970s, Smoothed Particle Hydrodynamics (SPH)~\citep{Lucy1977} is perhaps the most widely used mesh-free numerical method. It is commonly employed in a Lagrangian formulation where particles are advected with the underlying velocity field, where differential operators are approximated via radial kernel interpolation over neighbouring particles. SPH was originally developed for astrophysical problems~\citep{Lucy1977}, formulated with symmetries which guarantee exact global conservation, and it has since been applied more broadly to terrestrial fluid dynamics~\citep{Monaghan1994FreeSurfaceSPH}, in which settings global conservation is often sacrified in favour of anti-symmetric formulations exhibiting improved local accuracy. However, the method remains formally zero-th order consistent, limiting accuracy it can achieve, particularly in complex turbulent flows ~\citep{Liu2005,Quinlan2006}. This motivated the development of more accurate numerical methods that enforce polynomial consistency through the solution of a local or global linear system~\citep{Fornberg2011,Schrader2010,Gross2020,King2020}. However, consistent mesh-free methods are typically implemented in an Eulerian framework~\citep{Flyer2016,King2024}, as they require solving dense local linear systems for each local neighbourhood to obtain weights for each stencil. In a Lagrangian setting, where the particle configuration evolves in time, these methods require solving a local linear system for every particle at every time step, leading to substantial computational overhead. As a result, mesh-free simulations commonly face a trade-off between classical SPH kernels, which are computationally efficient but inconsistent and exhibit poor convergence, and consistency-corrected methods, which offer improved accuracy at the expense of significantly higher computational complexity.

In the current work, we introduce a self-supervised graph neural network framework that learns discrete \textit{mesh-free} differential operators directly from irregular particle configurations, named Neural Mesh-Free Differential Operator (NeMDO). Rather than learning PDE solutions or problem-specific closures, NeMDO constructs local operator weights by learning polynomial consistency constraints derived from Taylor expansions, yielding operators with a clear mathematical foundation. Our results show that graph neural networks can learn these consistency constraints and predict operator weights with structural properties analogous to those of classical mesh-free discretisations.

The primary objective of NeMDO is to demonstrate that neural networks can 
approximate polynomial consistency constraints for mesh-free discrete differential operators, and provide a flexible framework for constructing mesh-free operators whose accuracy and computational cost can be systematically controlled.

The resulting operators are strictly local, generalise across heterogeneous node configurations and resolutions, and are independent of any particular governing equation. We validate the learned operators using established numerical analysis tools, including convergence studies, modal response and stability analyses, and test the robustness of the framework through parametric studies. We further demonstrate applicability by solving the weakly compressible Navier–Stokes equations. The remainder of this paper is organized as follows. Section~\ref{sec:background} provides the theoretical background on discrete differential operators and machine learning for PDEs. In Section~\ref{sec:num_methods}, we detail traditional mesh-free operators, followed by a description of the proposed NeMDO methodology in Section~\ref{sec:method}. We then present a comprehensive series of experiments and discussions in Section~\ref{sec:experiments}. A parametric ablation study is provided in Section~\ref{app:ablation}, while Section~\ref{sec:limitations} highlights the limitations of the current framework. Finally, Section~\ref{sec:conclusion} offers concluding remarks and summarizes our key findings.

\section{Background \& Related Work}
\label{sec:background}
\subsection{Classical Numerical Approximation of Differential Operators}
~\label{sec:num_solver}
Given the discretisation of a spatial domain $\Omega$ by a set of collocation points
$\mathcal{P} := \{\mathbf{x}_i\}_{i=1}^N \subset \Omega \subset \mathbb{R}^d$,
a general local discrete approximation of a differential operator acting on $\phi$ can be written as
\begin{equation}
\label{eq:discrete_op}
    L^D(\phi(\mathbf{x}_i))=\sum_{j \in\ \mathcal{N}_i} (\phi(\mathbf{x}_j)-\phi(\mathbf{x}_i))w_{ji}^D,
\end{equation}
where $L^D$ denotes the discrete approximation of the differential operator $D$, $\mathcal{N}_i$ is the local support (or computational stencil) associated with node $i$, and $w_{ji}^D$ are the corresponding stencil weights. This formulation encompasses a broad class of numerical methods given appropriate stencils and indexing—including finite difference method~\citep{Lele1992}, finite element method~\citep{Clough1990}, SPH~\citep{Lucy1977}, and mesh-free high-order methods such as the Local Anisotropic Basis Function Method (LABFM)~\citep{King2020}—with the specific discretisation determined by how the weights $w_{ji}^D$ and neighbourhoods $\mathcal{N}_i$ are constructed.

The accuracy and utility of the discretised solution depends on the \textit{convergence, consistency}, and \textit{stability} of the numerical method. For a discretisation to be consistent, the discrete operator must reproduce the Taylor moments up to a prescribed order. In mesh-free methods, these consistency conditions are typically enforced by solving local reconstruction problems in each particle neighbourhood. Representative approaches include the reproducing kernel particle method~\citep{Liu1995}, generalised moving least squares (GMLS)~\citep{Trask2017,Trask2018}, the radial basis function-finite difference methods (RB-FDM)~\citep{tolstykh2000using,Bayona2013Flames,Bayona2015}, and LABFM~\citep{King2020}. While solving these local linear systems enforces polynomial consistency and can yield high-order convergence, these methods introduce high computational cost when computing stencil weights. This computational overhead can be prohibitive in Lagrangian/Semi-Lagrangian frameworks, when the particles move during time integration, and the stencil weights must be recomputed at every time step.

In SPH, the stencil weights are obtained with a compact isotropic kernel~\citep{Koschier2020}. The most used SPH kernel functions are the B-spline functions~\citep{Schoenberg1946} and the Wendland functions~\citep{Wendland1995}, where the error of the integral approximation is $\mathcal{O}(h^2)$ in the limit of large $h/s$ ~\citep{Monaghan2005}---where $h$ is the smoothing length of the kernel (i.e. the radius of the neighbouring region about point $i$) and $s$ is the average collocation point spacing---with convergence of $\mathcal{O}(h)$ or lower in practice~\citep{Vacondio2013,Ferrand2013,Fourtakas2019}. This discrepancy is largely attributed to the discretization of the original SPH formulation, which assumes a continuum; furthermore, irregularities in collocation point distribution can lead to further deterioration in accuracy~\citep{Quinlan2006}. A consistency analysis of discrete SPH indicates that gradient operators are zeroth-order accurate in $h$ on disordered particles, and certain Laplacian formulations may diverge as resolution increases. Nonetheless, convergence is often observed within a finite resolution window and for sufficiently well-behaved particle configurations and low wavenumbers~\citep{Quinlan2006}. The convergence behavior further depends on the neighbour count $\mathcal{N}_i$, which is kernel-dependent. In practice, achieving stable and convergent SPH discretisations typically requires large support sizes, often involving tens to hundreds of neighbours per particle~\cite{Dehnen2012}.

\subsection{Machine Learning for Partial Differential Equations}
~\label{sec:ml_pde}
\textbf{Phyics-Informed Neural Networks} is a prominent paradigm in scientific machine learning, where neural networks are trained to approximate the solution field associated with a specific PDE instance. This approach embed the governing equations, boundary conditions, and initial conditions into the training objective, typically by penalising PDE residuals evaluated through automatic differentiation~\cite{Raissi2019PINN,Gin2020DeepGreen,Karlbauer2021,Hao2023PIML}. This formulation enables mesh-free, unsupervised training, and can achieve high accuracy for well-posed problems. However, PINNs are inherently instance-specific, as the learned model corresponds to a particular choice of governing equations and initial/boundary conditions, typically requiring retraining when these change. 

\textbf{Neural Operators} constitute an alternative line of research that focuses on learning mappings between function spaces. Neural operator methods approximate the solution operator of a PDE from paired samples of input and output functions, enabling inference across varying discretisations and resolutions. Prominent examples include Deep Operator Networks (DeepONets)~\cite{Lu2021}, Fourier Neural Operators (FNOs)~\cite{Li2021FNO}, and graph-based neural operators (GNOs)~\cite{Anandkumar2019GKN}. These approaches learn implicit representations of the underlying PDE dynamics from data and, in some cases, have demonstrated transfer across related PDEs; however, the associated training and inference costs remain substantial~\cite{Subramanian2023Foundation}.

\textbf{Learning-based particle methods} leverage the locality of differential operators, to learn local mesh-free PDE time integration from solutions obtained from traditional operators~\citep{sanchezgonzalez2020,Li2022GNN,Toshev2023E3GNN,Toshev2024,Toshev2025QuasiLag}. While effective for specific flow classes, these approaches are typically trained end-to-end on time-dependent solution data, tying the learned representations closely to the governing equations and regimes observed during training.


\textbf{Learning Low-Level Numerical Operators: }A complementary line of work focuses on learning numerical discretisations or low-level operator components. Rather than approximating solutions directly, these approaches learn local numerical building blocks—such as derivative stencils or basis functions—while preserving the structure of established discretisation schemes. Examples include learning data-driven finite-difference stencils to enable resolution coarsening~\citep{BarSinai2019,Zhuang2021,Kochkov2021}, extending these to finite-volume methods~\citep{deRomemont2024}. While traditional neural operators and its variants can learn local differential operator actions from solution data~\citep{liuschiaffini2024neuraloperatorslocalizedintegral}, they do not explicitly construct reusable discrete operators as functions solely from stencil geometry. Despite these advancements, the ability of such models to approximate differential fields is often restricted to the spectral characteristics and functional forms present in the training distribution. 

More recently, \citet{Choi2025Foundation} proposed data-driven finite element methods (DD-FEM), which replaces classical polynomial bases with locally learned basis functions. This formulation enables reuse across geometries, mesh types, and boundary conditions, highlighting the potential of learning reusable numerical building blocks that remain compatible with classical discretisation principles. The framework presented in the present manuscript is closely aligned with this operator-level perspective, but targets the learning of mesh-free discrete differential operators rather than local basis functions in mesh-based discretisations.

\section{Standard Numerical Methods}
\label{sec:num_methods}
In this work, we use LABFM as an exemplar high-order mesh-free method, noting that it shares characteristics (structure of computational stencil, form of discrete operator, and local linear system) with other high-order mesh-free methods (e.g. generalised finite difference method~\citep{ZHENG20221}, GMLS~\citep{Trask2017}, RB-FDM~\citep{shankar2014radialbasisfunctionrbffinite}). For a comprehensive derivation of LABFM, we refer the reader to~\citep{King2020}. Additionally, we use two standard SPH kernels as benchmark: the quintic spline~\citep{Schoenberg1946} and the Wendland C2 kernel~\citep{Wendland1995}. Both have been widely adopted in computational fluid dynamics (CFD)-oriented simulations~\citep{Adami2013,Sun2018}. In the current section, we explain SPH and LABFM in detail.

\label{app:num_methods}
\subsection{Smoothed Particle Hydrodynamics Operators}
\label{app:sph}

For SPH gradient operators, one may choose the symmetric or anti-symmetric forms, with the former providing exact conservation and the latter providing greater accuracy. With the focus here on consistent operators, we consider only the anti-symmetric operator, which can be expressed in the form of Equation~\eqref{eq:discrete_op}, with
\begin{equation}
\label{eq:sph_grad}
    w_{ji}^\nabla = \nabla_i W(\mathbf{x}_{ji},h)V_j,
\end{equation}
where the subscript indicates $(\cdot)_{ji}= (\cdot)_j - (\cdot)_i$, thus, $\mathbf{x}_{ji}$ denotes the relative position of the collocation point $j$ with respect to point $i$, $W$ is a smoothing kernel and $V_j$ is the volume of the particle. When computing the Laplacian with SPH operators, we employ the Morris operator~\citep{Morris1997}:
\begin{equation}
    w_{ji}^\Delta =  \frac{-2\,\mathbf{x}_{ji}}{||\mathbf{x}_{ji} ||_2}\cdot\nabla_{i}W_{ji} V_{j}
\end{equation}
where $||\cdot||_2$ denotes the Euclidean distance.

\subsubsection{Smoothing Kernels}
\textbf{Quintic spline: } The quintic spline kernel has support $r \in [0,3]$ and its given by
\begin{equation}
    W^\text{QS}(r) = \sigma
    \begin{cases}
        ( 1 - r)^5_+ - 6\left(\frac{2}{3} - r \right )^5_+ + 15 \left (\frac{1}{3} - r\right )^5_+, & 0 \le r < 1, \\
        ( 1 - r)^5_+ - 6\left(\frac{2}{3} - r \right )^5_+ , & 1 \le r < 2, \\
        ( 1 - r)^5_+, & 2 \le r < 3, \\
        0, & r \ge 3,
    \end{cases}
\end{equation}
where $r=||\mathbf{x}_{ji}||_2/h$, $\sigma$ is a normalisation constant that depends on the system's dimension, in two dimensions $\sigma=\frac{7}{478\pi h^2}$~\citep{Liu2010SPH}. We set $h = 1.5s$, which yields approximately 60-65 neighbours for the quintic spline.

\textbf{Wendland C2: } The Wendland C2 kernel has a support of $r \in [0,2]$ and is defined as
\begin{equation}
    W^{\text{WC}2}(r) = \sigma (1 - r) ^ 4 (1 + 4r).
\end{equation}
In two dimensions, $\sigma=\frac{7}{\pi h^2}$. We set $h = 1.5s$, which yields approximately 25-30 neighbours for the Wendland C2.

\subsection{The Local Anisotropic Basis Function Method (LABFM)}
\label{app:labfm}

In LABFM, a general discrete operator is defined to approximate differential operators as shown in Equation~\eqref{eq:discrete_op}. The weights $w_{ji}^{D}$ are given by a weighted sum of anisotropic basis functions (ABFs),
\begin{equation}
w_{ji}^{D}=\mathbf{W_{ji}}\cdot\boldsymbol{\Psi_{i}^{D}}=\text{W}_{ji}^{1}\Psi_{i,1}^{D}+\text{W}_{ji}^{2}\Psi_{i,2}^{D}+\text{W}_{ji}^{3}\Psi_{i,3}^{D}+...
\end{equation}
in which the vector $\mathbf{W_{ji}}$ are the ABFs, and $\mathbf{\Psi^{D}_{i}}$ is a coefficient vector. The coefficient vector is obtained by solving the following linear system
\begin{equation}
\label{eq:linear_system}
        \mathbf{A_{i}}\boldsymbol{\Psi_{i}^{d}}=\mathbf{M^{D}}, 
\end{equation}
where $\mathbf{M^{D}}$ is a vector that contains the coefficients of the Taylor expansion about point $i$ in a computational stencil/support region, we will refer to this vector as the "moments" of the stencil. Below, we show instances of the moment vector when approximating exemplar differential operators
\begin{equation}
\label{eq:labfm_moments}
        \boldsymbol{C^{D}}=
        \begin{cases} 
            \text{$[1, 0, 0, 0, 0, 0, ...]^{T}$} & \text{$if$ $D=x$} \\
            \text{$[0, 1, 0, 0, 0, 0, ...]^{T}$} & \text{$if$ $D=y$} \\
            \text{$[0, 0, 1, 0, 1, 0, ...]^{T}$} & \text{$if$ $D=\Delta$}
        \end{cases} 
\end{equation}
For a given support region, $\mathbf{A_{i}}$ is given by
\begin{equation}
    \mathbf{A_{i}}=\sum_{j\in\mathcal{N}_i}\mathbf{X_{ji}}\otimes\mathbf{W_{ji}},
\end{equation}
where $\mathbf{X_{ji}}$ is defined as the vector of Taylor monomials,
\[
\mathbf{X}_{ji}
:=
\left[
x_{ji},
y_{ji},
\frac{x_{ji}^2}{2},
x_{ji}y_{ji},
\frac{y_{ji}^2}{2},
\frac{x_{ji}^3}{6},
\ldots
\right]^T,
\]
and the rank of $\mathbf{A_{i}}$ is given by, 
\begin{equation}
    \text{rank}(\mathbf{A_{i})}_{2D}=\frac{p^2+3p}{2} \quad \text{and} \quad  \text{rank}(\mathbf{A_{i})}_{3D}=\sum_{m=2}^{m=p+1}\frac{m(m+1)}{2}
\end{equation}
in 2 and 3 dimensions, respectively.

To construct the ABFs, bivariate Hermite polynomials are combined with a smoothing kernel. The $q$-th element of $\mathbf{X}_{ji}$ is proportional to $x_{ji}^{a}y_{ji}^{b}$. Thus, the $q$-th ABF is defined below in 2 dimensions
\begin{equation}
\label{eq:abf}
        W_{ji}^{q}=\frac{\psi\left(||\mathbf{x}_{ji}||_2/h_{i}\right)}{\sqrt{2^{a+b}}}H_{a}\left(\frac{x_{ji}}{h_{i}\sqrt{2}}\right)H_{b}\left(\frac{y_{ji}}{h_{i}\sqrt{2}}\right)
\end{equation}
where $H_{a}$ is the $a$-th order univariate Hermite polynomial (of the physicists kind), and $\psi$ is an RBF. In this work, the Wendland C2 kernel is used, following~\citet{King2022}.

\section{Method}
\label{sec:method}
In this section, we introduce NeMDO (Neural Mesh-Free Differential Operator), a new approach for computing weights used in discrete differential operators in disordered mesh-free simulations. Rather than computing these weights with an SPH kernel or by solving per-particle linear systems, we formulate operator construction as a learning problem defined over local particle neighbourhoods. We employ our framework to predict discrete operator weights based solely on the relative positions of neighbouring particles, enabling local mesh-free spatial differential approximations.

\textbf{Learning Mesh-Free Discrete Differential Operators. } Given a discretised domain with collocation points $\mathcal{P}$, we associate to each point $\mathbf{x}_i$ a local neighbourhood $\mathcal{N}_i:=\{\mathbf{x}_j \in \mathcal{P}: ||\mathbf{x}_{ji}||_2 \leq d_i^{(n)}\}$, where $d_i^{(n)}$ is the distance to the $n$-th nearest neighbor of $\mathbf{x}_i$. Our objective is to construct a local discrete approximation $L^D$ of a differential operator $D$ acting on $\phi$, such that the operator evaluation at $\textbf{x}_i$ is approximated by a weighted sum over neighbouring samples
\begin{equation}
\label{eq:gnn_op}
    L^D(\phi(\mathbf{x}_i))=\sum_{j \in \mathcal{N}_i} (\phi(\mathbf{x}_j)-\phi(\mathbf{x}_i))w_{ji}^D( \{\mathbf{x}_{ji}\}_{j \in \mathcal{N}_i} ;\theta), 
\end{equation}
where $w_{ji}^D(\{\mathbf{x}_{ji}\}_{j \in \mathcal{N}_i};\theta)$ are the data-driven weights inferred by a model parametrised by $\theta$, conditioned on the stencil geometry. For brevity, the explicit dependence on the stencil geometry is omitted in subsequent notation where it is understood from the context, i.e. $w_{ji}^D(\theta) \equiv w_{ji}^D(\{\mathbf{x}_{ji}\}_{j \in \mathcal{N}_i};\theta)$. We model the mapping from neighbourhood geometry to operator weights using a learnt function shared across all stencils. The learnt mapping does not depend on the field samples $\phi(\mathbf{x}_i)$, the governing equation, or the global simulation domain. Instead, it defines a reusable local operator constructor that can be applied consistently across particle distributions, resolutions, and governing equations.

\textbf{Local Operator Parametrisation:} For each local neighbourhood $\mathcal{N}_i$ a graph is defined $\mathcal{G}_i:=(\mathcal{V}_i,\mathcal{E}_i)$, where the node set $\mathcal{V}_i \equiv \mathcal{N}_i$ consists of the particle $i$ and its neighbours. The edge set $\mathcal{E}_i:=\{(i,j),(j,i) : j \in \mathcal{N}_i\setminus\{i\}\}$ defines star-shaped, bidirectional edges between the central particle and its neighbours. In contrast to dense radius graph seen in particle-based learning approaches~\citep{Li2022GNN}, this star-shaped construction yields linear complexity in the neighbourhood size, and the graph connectivity is directly constructed from the neighbourhood lists already computed in standard mesh-free methods, requiring no additional spatial searches or geometric preprocessing. Node attributes encode the normalised relative position
$\hat{\mathbf{x}}_{ji} := \mathbf{x}_{ji}/d_i^{(n)}$ ,
between neighbour particle $j$ and the central particle $i$.

The graphs $\mathcal{G}_i$ are processed by a shared parametric function $f_\theta$, implemented by a graph neural network (GNN),
\begin{equation}
    \{\hat{w}_{ji}^D(\theta)\}_{j \in \mathcal{N}_i}\leftarrow
    f_\theta \left(
    \mathcal{G}_i,
    \{\mathbf{\hat{x}}_{ji}\}_{j \in \mathcal{N}_i}
    \right).
\end{equation}
The relative positions are first embedded into a latent representation using a multi-layer perceptron (MLP), 
\begin{equation}
\label{eq:enc}
    \mathbf{v}_j^{0} 
    = \mathrm{MLP}^{\text{Emb}}_\theta\bigl(\hat{\mathbf{x}}_{ji}\bigr), \quad \mathrm{MLP}^{\text{Emb}}_\theta:\mathbb{R}^d \to \mathbb{R}^{F_h}, \quad \forall j\in \mathcal{N}_i,
\end{equation}
where $\mathbf{v}_j^{0}$ are the latent node features that are passed to the graph layers,  the number of dimensions of the system is given by $d$ and the number of hidden features per node after encoding is denoted as $F_h$. Message passing is then performed over the stencil graph, where we define $L$ graph layers as 
\begin{equation}
\label{eq:g_layer}
    \mathbf{v}_j^{l}
    = \mathrm{MLP}_\theta^{l,\text{Upd}}\Bigl(
        \mathbf{v}_j^{l-1},
        \,\oplus_{k \in \mathcal{K}_j} \; \mathrm{MLP}_\theta^{l,\text{Msg}}\left(\mathbf{v}_k^{l-1}\right)
      \Bigr),\quad l=1, ..., L, \quad\forall j\in \mathcal{N}_i,
\end{equation}
where $\mathrm{MLP}_\theta^{l,\text{Upd}}$ and $\mathrm{MLP}_\theta^{l,\text{Msg}}$ are the update and message networks at layer $l$, $\mathbf{v}_j^{l}$ denotes the node features of node $j$ at graph layer $l$, $\mathcal{K}_j\subset\mathcal{N}_i$ is the set of nodes connected to node $j$, and $\oplus$ is a differentiable, permutation-invariant aggregation operator. In our implementation, $\oplus$ is realised as an attention-weighted aggregation based on cross-graph matching~\citep{Li2019GraphMatching}.

\begin{figure}[ht]
  \centering 
  \includegraphics[width=0.21\linewidth, angle=-90]{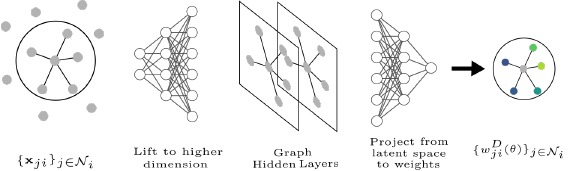}
  \caption{We learn the mapping from relative position of particles within a local neighbourhood, to a local set of weights that approximate differential operators. The learned operator is physics-agnostic, and local to the computational stencil. The framework consists of a lifting the relative positions to a higher dimension with a neural network, followed by a stack of message-passing graph layers, and a final output head that maps latent representations to operator weights.}
  \label{fig:flow_diagram}
\end{figure}

Lastly, a final MLP maps the latent node representations to normalised discrete operator weights,
\begin{equation}
\label{eq:dec}
    \hat{w}_{ji}
    = \mathrm{MLP}^{\text{Out}}_\theta(\mathbf{v}_j^{L}), \quad \mathrm{MLP}^{\text{Out}}_\theta:\mathbb{R}^{F_h} \to \mathbb{R}, \quad \forall j\in \mathcal{N}_i
\end{equation}
in which $\{ \hat{w}_{ji}(\theta) \}_{j \in \mathcal{N}_i}$ are the weights associated with the target differential operator. A schematic of the framework can be seen in Figure~\ref{fig:flow_diagram}. All MLPs use hyperbolic tangent activation functions. Models are trained using the adaptive moment estimation (Adam) optimiser. In the present work, we focus on demonstrating the potential of this form of method, and reserve an assessment of the effect on learning of different optimisers for future work.

Translation invariance is enforced by encoding geometry through relative position vectors, while permutation invariance is achieved through symmetric aggregation operations over neighbouring nodes (which is inherently handled by the graph architecture). Scale robustness across particle resolutions is introduced by normalizing relative distances with respect to the maximum distance between neighbouring particles and the central particle. The proposed architecture is not explicitly rotation-equivariant; instead, rotational robustness is obtained implicitly through training on neighbourhoods with diverse node configurations.

\textbf{Self-Supervised Learning with Polynomial Consistency Constraints:}
To learn the parameters $\theta$, we employ a training objective derived from polynomial consistency conditions imposed by a truncated Taylor expansion. Consequently, the method does not require labelled target weights for individual stencils. To formulate the loss function, let us first consider a scalar field in 2 dimensions $\phi: \mathbb{R}^2 \to \mathbb{R}$ sampled on a discrete point cloud $\mathcal{P}$. For a given neighbourhood $\mathcal{N}_i$, where $\mathbf{D}(\phi)|_i$ denotes the vector of partial derivatives of $\phi$ at particle $i$, 

\[\mathbf{D}(\phi)|_i:=\left [ \left.\frac{\partial\phi}{\partial x}\right |_i, 
\left.\frac{\partial \phi}{\partial y}\right |_i, 
\left.\frac{\partial^2\phi}{\partial x^2}\right |_i, 
\left.\frac{\partial^2 \phi}{\partial x \partial y }\right |_i, 
\left.\frac{\partial ^2\phi}{\partial y^2}\right |_i, 
\left.\frac{\partial ^3 \phi}{\partial x^3}\right |_i, ... \right ]^T,\] and $\mathbf{X}_{ji}$ is the vector of Taylor monomials of the position of $j$ relative to $i$, the multivariate Taylor expansion of $\phi$ about point $i$ may be written compactly as 
\begin{equation}
    (\phi)_j=(\phi)_i \;+ \mathbf{X}_{ji} \:\cdot \mathbf{D}(\phi)|_i
\end{equation}

One can analyse the error in the discrete differential operator $L^D$ by substituting the Taylor expansion into Equation~\eqref{eq:discrete_op}, obtaining
\begin{equation}
    L_i^D(\phi)=\sum_{j \in \mathcal{N}_i}\mathbf{X}_{ji} \:\cdot \mathbf{D}(\phi)|_i \,w_{ji}^D
\end{equation}

For a target continuous differential operator $D$, polynomial consistency of order $p$ requires that the discrete operator $L_i^D$ reproduces the action of $D$ exactly on all monomials of total degree at most $p$, i.e.
\begin{equation}
    L_i^D (\phi) = D(\phi)|_i, \quad \forall \;\text{polynomials of degree} \leq p.
\end{equation}
For instance, to approximate $\frac{\partial\phi}{\partial x}$ with second-order polynomial consistency, it is required that the first-order discrete moments satisfy
\begin{equation}
        \sum_{j \in \mathcal{N}_i}x_{ji}w_{ji}^x=1,\quad  \sum_{j \in \mathcal{N}_i}y_{ji}w_{ji}^x=0
\end{equation}
with all remaining moments associated with the second-order approximation set to 0. When these conditions are satisfied, the resulting discrete operator approximates $\frac{\partial\phi}{\partial x}$ with second-order convergence for sufficiently smooth $\phi$, yielding a truncation error of order $s^2$.


In our framework, we approximate polynomial consistency by minimizing the error between predicted and target moments over all particles
\begin{equation}
\mathcal{L}
=
\frac{1}{N}
\sum_{i=1}^N
\left\|
\sum_{j \in \mathcal{N}_i}  \hat{\mathbf{X}}_{ji}\,\hat{w}_{ji}^D(\theta) - \mathbf{M}^D
\right\|^2_2,
\end{equation}
where $\hat{}$ indicates a normalised quantity. Thus, the proposed method does not require individual labels for each computational stencil. One can compute the training operator moments directly from the inputs and the predicted weights, and compute the loss based on the fixed target moments. 




\textbf{Data Generation:}
Training is performed on local particle neighbourhoods $\mathcal{N}_i$ sampled from synthetic disordered point cloud distributions. These are obtained by applying stochastic perturbations to a regular Cartesian grid with average spacing $s_u$. Specifically, each grid node is independently displaced by a noise term sampled from a uniform distribution $\mathcal{U}(-\frac{\epsilon s_u}{2}, \frac{\epsilon s_u}{2})$ per coordinate, where $\epsilon$ represents the non-dimensional noise intensity. This formulation ensures that the perturbation magnitude scales proportionally with the grid resolution, enabling a systematic analysis of operator sensitivity across different discretization scales.

Unless stated otherwise, models are trained at a disturbance level of $\epsilon = 1.0$, which is significantly larger than the particle disorder typically encountered in practical mesh-free simulations. This choice ensures that the learned operators remain robust under severe geometric irregularity.

\textbf{Operator Rescaling and Reuse:} For a target differential operator of order $m$, the predicted weights are rescaled by $(d_i^{(n)})^{-m}$ to recover the correct physical dimensions, reflecting the $m$-th order spatial scaling of the operator---distinct from the formal order of accuracy $p$---and ensuring consistency across resolutions. 
The resulting weights are then used within standard mesh-free discretisations to approximate the action of the operator $D$ on arbitrary fields.

Because operator construction depends only on local geometry, trained models can be reused across different resolutions, domains, and governing equations without retraining, provided the neighbourhood size $|\mathcal{N}_i|$ remains fixed. This enables learned operators to be deployed as drop-in numerical components within existing mesh-free solvers.

\section{Results \& Discussion}
\label{sec:experiments}
The present framework is assessed using a combination of qualitative analysis, polynomial reproduction residuals, derivative error analysis on a test function, modal response, stability analysis, parametric investigation and behaviour in fluid simulations. To assess the computational efficiency of the proposed framework, we report a wall-clock time analysis as a function of the $L_2$ error on a test function. Unless otherwise stated, all results are reported for a canonical configuration consisting of a first-order derivative or Laplacian operators, both with with second-order consistency. Details on hyper-parameters and the training dataset for each specific model can be found in Appendix~\ref{app:hyper_p}. 

\subsection{Qualitative Analysis of Learned Operator}
\label{app:qual_analysis}
\begin{figure}[h]
\begin{center}
  \begin{subfigure}{0.3\linewidth}
    \begin{center}
      \includegraphics[width=\linewidth]{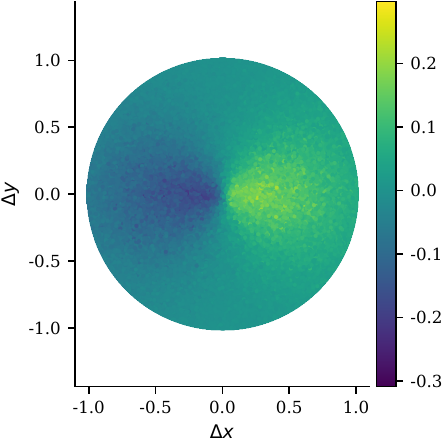}
    \end{center}
    \caption{$x$-derivative}
    \label{fig:kernelA}
  \end{subfigure}
  \hspace{0.04\linewidth}
  \begin{subfigure}{0.3\linewidth}
    \begin{center}
      \includegraphics[width=\linewidth]{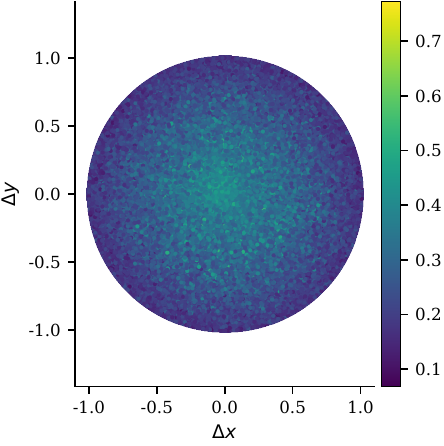}
    \end{center}
    \caption{Laplacian}
    \label{fig:kernelB}
  \end{subfigure}
\end{center}
\caption{Second-order normalised learned operators, colours indicate weight value. Models were trained with particle disturbance of $\epsilon=1.0$ and inferred with the same noise.}
\label{fig:kernels}
\end{figure}

To assess the geometric structure of the learned discrete operators, we perform a qualitative analysis of the predicted stencil weights across multiple noisy neighbourhood realisations. Figure~\ref{fig:kernels} visualizes the learned kernel corresponding to two differential operators, where weights predicted for many independently perturbed neighbourhoods are overlaid in relative coordinates.

Despite significant stochastic variation in local particle configurations, the learned operators exhibit coherent structures. In Figure~\ref{fig:kernels}, the $x$-derivative aggregated weight distribution displays approximate anti-symmetry in the direction of the derivative about the origin, i.e. $w(\mathbf{x}_{ji}) \approx -w(-\mathbf{x}_{ji})$, consistent with the structure of a first-derivative operator, while decaying in the transverse direction. This symmetry is not enforced explicitly but emerges purely from the polynomial moment constraints used during training. The overlaid kernels further reveal a decay in weight magnitude with increasing distance from the central particle. For Laplacian operator, the learned kernels instead exhibit rotational invariance, with larger weights concentrated near the central particle and decreasing with distance. Together, these symmetry patterns and the consistent decay of weight magnitude away from the central particle mirror structural properties commonly observed in gradient and Laplacian kernels of classical mesh-free methods, including uncorrected SPH and polynomially corrected kernel formulations~\citep{Dehnen2012, King2020}. 

\subsection{Polynomial Consistency and Convergence} 
\label{sec:convergence}
We now verify that the learned operators satisfy the polynomial consistency conditions imposed in the training loss and that this translates into accurate differential approximations. For each operator type, we evaluate the learned stencil weights on monomials up to degree \(p\). For brevity, we only show the first derivative results for the $x$ direction, noting that results in the $y$ direction are analogous. 

\begin{table}[t]
\caption{Moment residuals for learned NeMDO and SPH operators. Mean absolute error (MAE) and standard deviation are reported for each monomial moment, averaged over independently perturbed neighbourhood realisations. Residuals quantify the extent to which the learned discrete operators satisfy the imposed Taylor consistency constraints (smaller values are better). The superscripts in the learned operator identify the target operator.}
\label{tab:moments-models}
\begin{center}
\setlength{\tabcolsep}{8pt} 
\renewcommand{\arraystretch}{1.1} 
\begin{tabular}{ccccccc}
\multicolumn{1}{c}{\bf Operator} &
\multicolumn{1}{c}{\bf Metric} &
\multicolumn{1}{c}{\bf $x$} &
\multicolumn{1}{c}{\bf $y$} &
\multicolumn{1}{c}{\bf $x^2/2$} &
\multicolumn{1}{c}{\bf $xy$} &
\multicolumn{1}{c}{\bf $y^2/2$}
\\ \hline \\

\multirow{2}{*}{$\text{NeMDO}_{p=2}^x$}& MAE   & $8.84 \times 10^{-5}$ & $7.93 \times 10^{-5}$ & $5.77 \times 10^{-5}$ & $5.59 \times 10^{-5}$ & $4.74 \times 10^{-5}$ \\
                      & St.d. & $1.14 \times 10^{-4}$ & $1.04 \times 10^{-4}$ & $7.38 \times 10^{-5}$ & $7.35 \times 10^{-5}$ & $6.25 \times 10^{-5}$ \\
\hline
\multirow{2}{*}{$\text{NeMDO}_{p=2}^\Delta$} & MAE   & $5.22 \times 10^{-4}$ & $5.09 \times 10^{-4}$ & $3.69 \times 10^{-4}$ & $4.09 \times 10^{-4}$ & $3.62 \times 10^{-4}$ \\
                           & St.d. & $6.91 \times 10^{-4}$ & $6.69 \times 10^{-4}$ & $4.94 \times 10^{-4}$ & $5.49 \times 10^{-4}$ & $4.77 \times 10^{-4}$ \\
\hline

\multirow{2}{*}{$\text{Wendland C2}^x$} & MAE   & $5.14 \times 10^{-2}$ & $4.01 \times 10^{-2}$ & $2.70 \times 10^{-2}$ & $2.63 \times 10^{-2}$ & $1.36 \times 10^{-2}$ \\
                           & St.d. & $6.21 \times 10^{-2}$ & $5.01 \times 10^{-2}$ & $3.35 \times 10^{-2}$ & $3.29 \times 10^{-2}$ & $1.71 \times 10^{-2}$ \\
                           \hline
\multirow{2}{*}{$\text{Wendland C2}^\Delta$} & MAE   & $1.86 \times 10^{-1}$ & $1.83 \times 10^{-1}$ & $5.14 \times 10^{-2}$ & $8.02 \times 10^{-2}$ & $5.26 \times 10^{-2}$ \\
                           & St.d. & $2.31 \times 10^{-1}$ & $2.27 \times 10^{-1}$ & $6.21 \times 10^{-2}$ & $1.00 \times 10^{-1}$ & $6.40 \times 10^{-2}$ \\
\hline
\multirow{2}{*}{$\text{Quintic Spline}^x$} & MAE   & $3.44 \times 10^{-2}$ & $2.42 \times 10^{-2}$ & $1.98 \times 10^{-2}$ & $2.09 \times 10^{-2}$ & $1.10 \times 10^{-2}$ \\
                           & St.d. & $4.24 \times 10^{-2}$ & $3.03 \times 10^{-2}$ & $2.46 \times 10^{-2}$ & $2.59 \times 10^{-2}$ & $1.38 \times 10^{-2}$ \\
                           \hline
\multirow{2}{*}{$\text{Quintic Spline}^\Delta$} & MAE   & $8.49 \times 10^{-2}$ & $7.99 \times 10^{-2}$ & $3.44 \times 10^{-2}$ & $4.85 \times 10^{-2}$ & $3.35 \times 10^{-2}$ \\
                           & St.d. & $1.05 \times 10^{-1}$ & $9.91 \times 10^{-2}$ & $4.24 \times 10^{-2}$ & $6.05 \times 10^{-2}$ & $4.15 \times 10^{-2}$ \\

\end{tabular}
\end{center}
\end{table}

Table~\ref{tab:moments-models} reports moment residuals for learned NeMDO and SPH operators, averaged over independently perturbed neighbourhood realisations. For the gradient operator, NeMDO residuals are consistently on the order of \(10^{-5}\) across all first- and second-order monomials, while SPH residuals are of order \(10^{-2}\). For all operator, the Laplacian exhibits larger residuals for all operators compared to the derivative ones, with NeMDO on the order of \(10^{-4}\), and the SPH operators with errors of order \(10^{-1} - 10^{-2}\). The larger residuals for the Laplacian reflects the increased sensitivity of second-order derivatives to geometric perturbations, a behaviour also observed in classical high-order mesh-free and spectral discretisations~\citep{Lin2025ISPH}. Overall, these results indicate that the polynomial consistency constraints can be learnt with a self-supervised framework, generalise for unseen stencil arrangements, and significantly outperform SPH operators.

Next, we evaluate the learned operators on a smooth test function with known analytical derivatives. We considered a square domain defined by $(x,y) \in [-0.5,0.5]^2$, and we define the test function as
\begin{equation}
\label{eq:test_func}
    \phi(\tilde{x},\tilde{y})=1.0 + (\tilde{x}\tilde{y})^4+\sum_{n=1}^6(\tilde{x}^n+\tilde{y}^n),
\end{equation}
where $\tilde{x}=x-0.1453$ and $\tilde{y}=y-0.16401$. This choice of test function with pseudo-random offset ensures asymmetry in the function, to prevent the masking of errors, which could cancel for a symmetric function (e.g. Fourier modes)~\citep{King2020}.

Relative $L_2$ errors between the predicted and exact derivatives are reported at varying average particle spacing $s$. The learned operators are compared against classical discretisations, including SPH kernels based with the quintic spline and Wendland C2, as well as a representative formally consistent method, LABFM. 

\begin{figure}[t]
  \begin{center}
    \begin{subfigure}{0.35\linewidth}
      \centering
      \includegraphics[width=\linewidth]{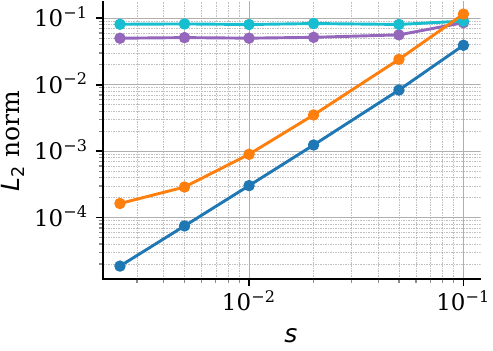}
    \end{subfigure}
    \begin{subfigure}{0.54\linewidth}
      \centering
      \includegraphics[width=\linewidth]{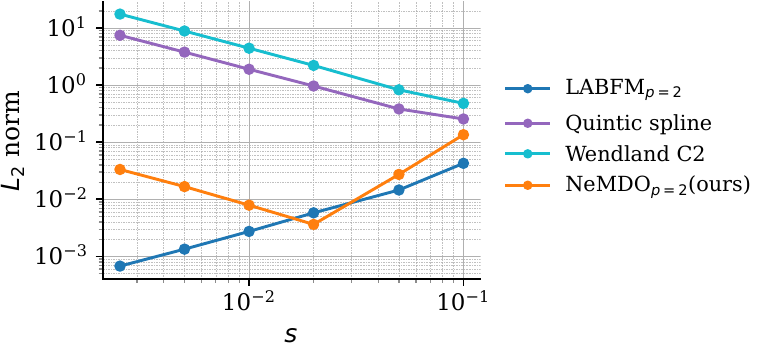}
    \end{subfigure}
  \end{center}
  \caption{Convergence of the discrete $x$-derivative (\textit{left}) and Laplacian (\textit{right}) on a smooth test function with $\epsilon = 0.5$, noting that NeMDO was trained with $\epsilon = 1.0$. Relative $L_2$ error versus particle spacing $s$ for the learned NeMDO operator, a second-order LABFM operator, and two uncorrected SPH kernels (quintic spline and Wendland C2).}
  \label{fig:convergence}
\end{figure}

Figure~\ref{fig:convergence} reports the convergence behaviour of the derivative and Laplacian. All convergence results are computed at a particle disorder level of $\epsilon = 0.5$ given SPH convergence is known to degrade significantly at higher disorder levels~\citep{Quinlan2006}. For the gradient, $\text{NeMDO}_{p=2}$ closely follows the second-order LABFM$_{p=2}$ scheme, exhibiting a decay rate consistent with second-order convergence as $s \to 0$. Across all resolutions, both the learned and LABFM operators substantially outperform the SPH kernel. For most mesh-free operators (including SPH and LABFM), achieving the expected convergence behaviour requires a sufficiently smooth underlying function, moderate particle disorder, and an adequate support size. Under these conditions, the limiting accuracy in consistent methods is typically bounded by the accuracy of the solution of the linear system~\citep{shankar2014radialbasisfunctionrbffinite}. NeMDO is not a formally consistent method (polynomial consistency is learnt, not enforced); thus, the observed error floor is instead governed by the residual moment errors and standard deviation reported in Table~\ref{tab:moments-models}. The comparatively poor SPH performance is attributed to the high particle disorder and the complexity of the test function, both of which are known to degrade uncorrected kernel approximations~\cite{Quinlan2006}.

For the Laplacian, LABFM exhibits the expected first-order convergence, achieving the lowest errors over the range of resolutions considered. The learned Laplacian  displays an apparent second-order decay at coarse resolutions; however, this behaviour is only observed over a small range of resolutions, and while first-order convergence is expected, we observe second-order convergence because higher-order truncation error terms dominate the error when $s$ is large (this effect is strongly dependent on the choice of $\phi$). As resolution increases, the error saturates at a limiting value of order $10^{-3}$, after which the error grows with order $s^{-1}$, similarly to traditional SPH formulations~\citep{Meng2025HighOrderSPH}. This behaviour is consistent with the presence of a residual consistency error in the Laplacian operator, with similar divergence observed in formally consistent mesh-free methods and spectral SPH formulations, albeit at finer resolutions~\citep{King2022,Lin2025ISPH}. Despite this limitation, the learned Laplacian remains significantly more accurate than uncorrected SPH kernels across the entire range of $s$. 

Overall, these results suggest that learning to approximate polynomial consistency provides meaningful improvements over traditional SPH and exhibits similar behaviour to the formally consistent method within a finite resolution range.

\subsection{Stability} 
\label{sec:stability}
To analyse the stability of the discrete operators, we construct a global discrete derivative matrix $\mathbf{G}^D$ as a re-arrangement of the local operators $L_{i}^{D}$. The global discrete operator matrix acts as a mapping that describes how a global field $\phi$ evolves across the \textit{entire domain} simultaneously, and the eigenvalues of $\mathbf{G}^D$ provide insight into the stability properties of the discretisation. Eigenvalues on the imaginary axis correspond to advective (translational) modes, while eigenvalues with nonzero real parts indicate growth or decay. These are of particular interest in dynamical systems, where the spectral distribution of $\mathbf{G}^D$ dictates the long-term behavior of the numerical solution.

For convective derivatives (i.e. first-order spatial derivatives), the continuous operator is purely dispersive, with Fourier modes corresponding to translation and no amplification or attenuation. Accordingly, the eigenvalues $\mu$ of a stable discrete approximation should ideally lie on the imaginary axis, i.e. $\Re(\mu)=0$ $\forall\;\mu$. In contrast, the Laplacian operator is purely dissipative, i.e. $\Re(\mu)\leq 0 \; \; \forall\;\mu$, reflecting the decay of all modes due to diffusion~\citep{Fornberg2011}.

\begin{figure}[h]
  \begin{center}
    \begin{subfigure}{0.36\linewidth}
      \centering
      \includegraphics[width=\linewidth]{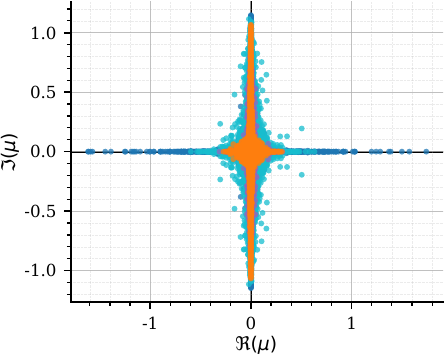}
    \end{subfigure}
    \begin{subfigure}{0.59\linewidth}
      \centering
      \includegraphics[width=\linewidth]{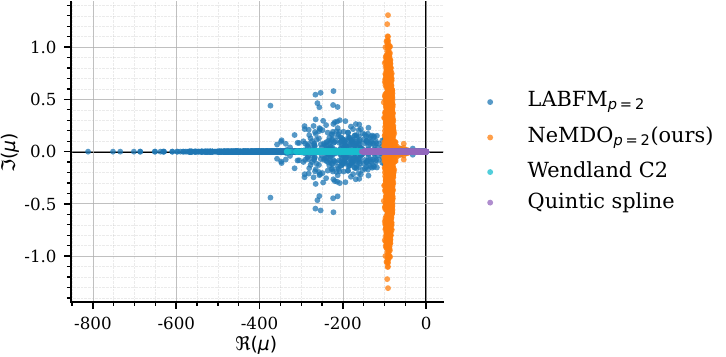}
    \end{subfigure}
  \end{center}
  \caption{Normalised eigenvalue spectrum of the global $x$-derivative operator with node disturbance $\epsilon = 1.0$ and 2500 nodes. \textit{Left}: $x$-derivative operators; \textit{right}: Laplacian operators.}
  \label{fig:stability}
\end{figure}

Figure~\ref{fig:stability} (\textit{left}) shows the normalised eigenvalues of the discrete $x$-derivative operator on a noisy particle distribution. The spectrum of the learned operator is tightly clustered near the imaginary axis, with a comparable spread to the quintic spline, and lies closer to the imaginary axis than that of LABFM constructed with the same neighbourhood size $|\mathcal{N}_i|$ and the Wendland C2 kernel. Given that high-order discretisations are known to generate small-scale oscillations~\citep{Lamballais2011}, it is notable that the learned operator—trained to approximate second-order polynomial consistency—exhibits stability properties comparable to the quintic spline. This suggests that the learned operator achieves more favourable stability characteristics per neighbour than both corrected and uncorrected kernel-based baselines.

For the Laplacian, no eigenmodes exhibit growth in time ($\Re(\mu)\le 0\;\forall\,\mu$) for any of the operators considered. The Morris Laplacian~\citep{Morris1997} exhibits only real components, corresponding to purely diffusive behaviour, which likely contributes to its widespread use and robustness in SPH simulations. NeMDO's Laplacian displays a larger spread along the imaginary axis than the other methods; however, its eigenvalues are more tightly clustered overall, indicating more uniform damping of modes. Overall, the learned operators demonstrate a more uniform spectral response across modes in disordered particle configurations for both first- and second-order derivatives, although a large dispersiveness is observed in the Laplacian. It should be noted that the results presented in this section are with particle disorder of $\epsilon =1.0$, which represents the worst-case scenario regarding node distribution compared to standard simulations that traditionally use particle shifting~\citep{Xu2009ISPH,Lind2012ISPH, Oger2016}.

\subsection{Modal Response}
\label{sec:modal_resp}
Insight into the behaviour of the derivative operators can be gained from an analysis of the modal response~\citep{broadley2025}. In principle, spectral discretisations can exactly represent all modes below the Nyquist limit~\cite{Brandenburg2003}---where the Nyquist wavenumber is given by $k_{Ny}=\pi/s_i$---however, truncated discretisations only approximate the underlying function up to a finite truncation error, leading to a modified (effective) wavenumber response. 

The modal response can be analysed by setting $\phi=\phi(\mathbf{x}_{ji})$ to be a local (to each stencil) multi-dimensional Fourier series; in our case, we restrict the analysis to a two-dimensional domain
\begin{equation}
    \phi(x,y) = \sum_{k_x \in \mathbb{Z}}\:\sum_{k_y \in \mathbb{Z}} c_{k_x,k_y}e^{i(k_x x+ k_yy)},
\end{equation}
where the $c_{k_x,k_y}$ are complex constants and $k_x$, $k_y$ are (appropriately scaled) wavenumbers. The effective wavenumber is then $k_{eff}=L^D_i(\phi)$ representing the wavenumber captured by the discrete operator. For the first derivative with respect to $x$, the effective wavenumber in the $i$-th stencil is given by
\begin{equation}
\label{eq:grad_modal}
    k_{eff} = \sum_{j\in \mathcal{N}_i} \sin (k_x x_{ji} + k_y y_{ji}) w_{ji}^x +i\sum_{j \in \mathcal{N}_i} (1- \cos (k_x x_{ji} + k_y y_{ji})) w_{ji}^x.
\end{equation}
The real part of Equatiom~\eqref{eq:grad_modal} characterizes the numerical scheme's dispersion error; specifically, an error exists whenever $\Re\{k_{eff}\} \neq k_x$. Similarly, the imaginary part represents dissipation error, occurring if $\Im\{k_{eff}\} \neq 0$. For the Laplacian, the effective wavenumber is given by
\begin{equation}
    q_{eff}^2 = \sum_{j\in \mathcal{N}_i}(1-\cos(k_x x_{ji} + k_y y_{ji}))w_{ji}^\Delta - i\sum_{j\in \mathcal{N}_i}\sin (k_x x_{ji} +k_y y_{ji})w_{ji}^\Delta
\end{equation}
where $q^2_{eff} = k_x^2 + k_y^2$ for spectral methods. In the equation above, the real part of $q_{eff}^2$ relates to the dissipation, and the imaginary part to the dispersion error.

\begin{figure}[h]
  \begin{center}
    \begin{subfigure}{0.44\linewidth}
      \centering
      \includegraphics[width=\linewidth]{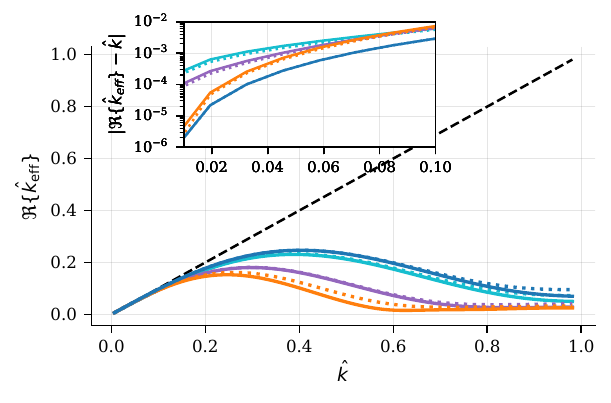}
    \end{subfigure}
    \begin{subfigure}{0.54\linewidth}
      \centering
      \includegraphics[width=\linewidth]{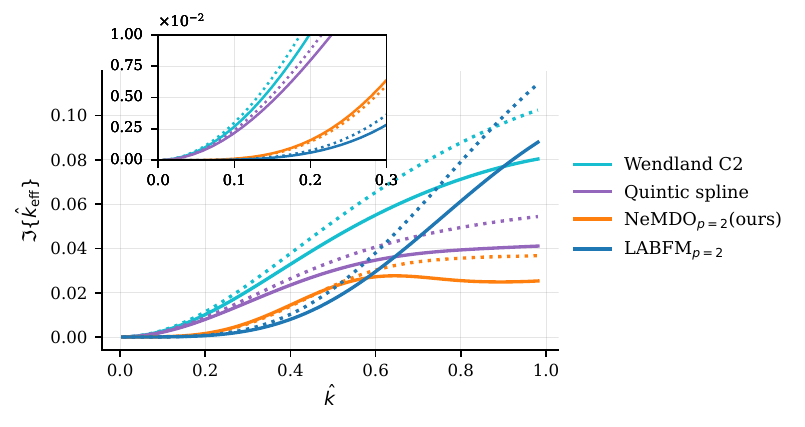}
    \end{subfigure}
  \end{center}
  \caption{Resolving power of the $x-$derivative for LABFM, NeMDO, and SPH (with the Quintic Spline and Wendland C2). The \textit{left} plot shows the real part of the modal response, and the \textit{right} plot shows the imaginary part on disordered particle distributions with noise level $\epsilon$. The horizontal axis shows the true wavenumber $k$ normalised by the Nyquist wavenumber $k_{\mathrm{Ny}}$, while the vertical axis shows the corresponding effective wavenumber $k_{\mathrm{eff}}/k_{\mathrm{Ny}}$. The dashed black line indicates the ideal response, corresponding to a spectral method. The inset on the \textit{left} plot shows the absolute difference between the different operators and the ideal spectral response. The inset on the \textit{right} plot shows zoomed in graph of $\Im\{\hat{k}_{eff}\}$ for range $0\leq\hat{k}\leq 0.3$. The full and dotted lines indicate $k_y/k_x=0$ and $k_y/k_x=1$, respectively.}
  \label{fig:res_power_re}
\end{figure}

\begin{figure}[h]
  \begin{center}
    \begin{subfigure}{0.44\linewidth}
      \centering
      \includegraphics[width=\linewidth]{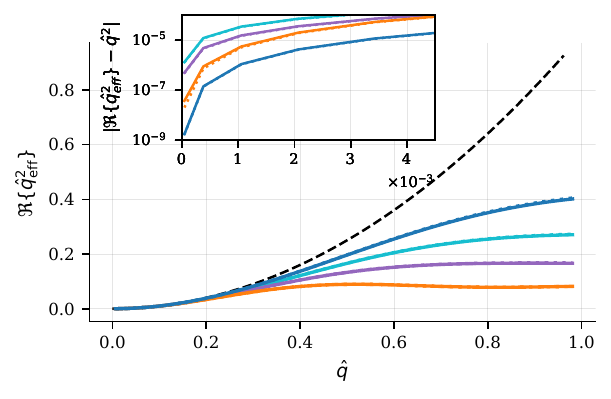}
    \end{subfigure}
    \begin{subfigure}{0.54\linewidth}
      \centering
      \includegraphics[width=\linewidth]{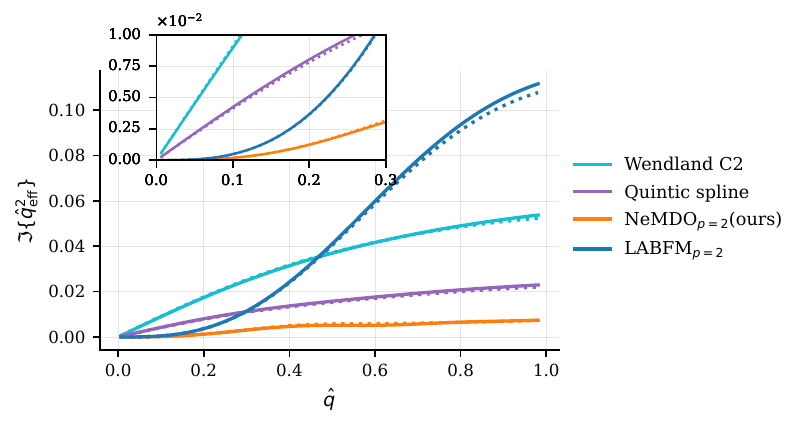}
    \end{subfigure}
  \end{center}
  \caption{Resolving power of the Laplacian operator, the \textit{left} plot shows the real part of the modal response and the \textit{right} plot shows the imaginary part on disordered particle distributions with noise level $\epsilon = 1.0$. The horizontal axis shows the true wavenumber $q$ normalised by the Nyquist wavenumber $k_{\mathrm{Ny}}$, while the vertical axis shows the corresponding effective wavenumber. The dashed black line indicates the ideal response, corresponding to a spectral method. The inset on the \textit{left} plot shows the absolute difference between the different operators and the ideal spectral response. The inset on the \textit{right} plot shows zoomed in graph of $\Im\{\hat{q}_{eff}^2\}$ for range $0\leq\hat{q}\leq 0.3$. The full and dotted lines indicate $k_y/k_x=0$ and $k_y/k_x=1$, respectively.}
  \label{fig:res_power_im}
\end{figure}

Figure~\ref{fig:res_power_re} shows the modal response of the different discretisations for the $x-$derivative operator, and Figure~\ref{fig:res_power_im} for the Laplacian. For both differential operators, all methods recover the true modal response at low wavenumbers and progressively deviate as the true wavenumber increases, indicating larger errors for high wavenumber modes. In both cases, the real part of LABFM$_{p=2}$ remains closest to the spectral reference, followed by the Wendland C2 and quintic spline kernels, while the learnt operator exhibits the lowest resolving power for the majority of wavenumbers. However, the insets in Figures~\ref{fig:res_power_re} (\textit{left}) and ~\ref{fig:res_power_im} (\textit{left}) show that at low wavenumbers, NeMDO outperforms SPH by approximately two orders of magnitude. Although the SPH kernels exhibit a stronger modal response than NeMDO for higher wavenumber, this spectral advantage is confined to sufficiently low resolutions and did not translate into smaller error during the convergence study in practice due to their pronounced sensitivity to node disorder shown in Figure~\ref{fig:convergence} and the relatively fine resolutions plotted. The imaginary parts of the differential operators indicate that NeMDO has smaller dissipation and dispersion errors in the $x-$derivative and the Laplacian, respectively, compared to the other operators. 

The global spectral analysis (Section~\ref{sec:stability}) and the local modal response (Section~\ref{sec:modal_resp}) provide complementary perspectives on the numerical performance of these operators regarding dispersion and dissipation. While both analyses reflect the influence of particle irregularity, they do so at different scales of aggregation. The modal response characterizes the local fidelity of the operator, demonstrating that NeMDO weights are effectively optimized to suppress dispersion error on a per-stencil basis, even under high disorder. Conversely, the global spectrum captures the matrix non-normality of the assembled system. The large imaginary values in the NeMDO Laplacian (Figure~\ref{fig:stability}) are not a result of local dispersive failure—--which the modal response proves is minimal--—but rather a manifestation of the global matrix non-normality. This arises from the lack of reciprocity between neighbors in a disordered mesh-free assembly ($w_{ij} \neq w_{ji}$).

\subsection{Computational Cost and Accuracy }
We now evaluate the cost–accuracy trade-off of NeMDO relative to LABFM and SPH. For each configuration, we measure the wall-clock time required to compute stencil weights for the $x$-derivative on a fixed particle cloud and evaluate the resulting $L_2$ error of the test function in Equation~\eqref{eq:test_func}, further details regarding the computational cost can be found in Appendix~\ref{app:comp_cost}. We report results for NeMDO$_{p=2}^1$, NeMDO$_{p=2}^2$, NeMDO$_{p=2}^3$, where the superscripts identifies increasing numbers of trainable parameters, to examine how computational cost and limiting accuracy scale with model capacity.

\begin{figure}[h]
  \begin{center}
    \includegraphics[width=0.5\linewidth]{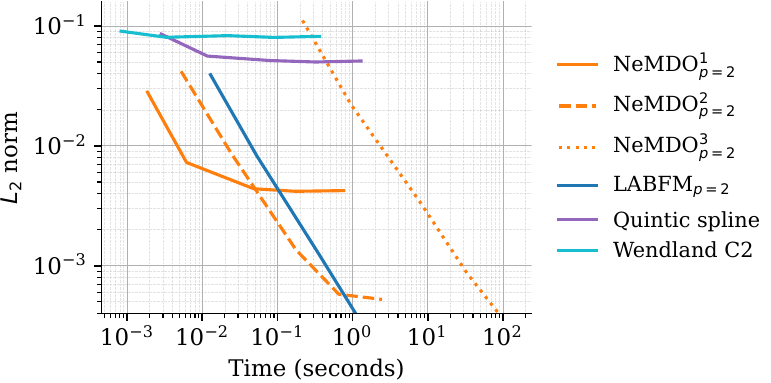}
  \end{center}
  \caption{Evaluation of the trade-off between parameter count, wall-clock forward time, and accuracy for NeMDO and baseline kernel methods when computing stencil weights. The vertical axis shows the $L_2$ error of the $x$-derivative of the test function in Equation~\eqref{eq:test_func} at a different resolutions; horizontal lines mark the resolution at which each operator converges (i.e., further refinement does not reduce this error). All NeMDO models are trained with geometric noise $\epsilon = 1.0$, while the convergence tests are performed at $\epsilon = 0.5$. }
  \label{fig:error_time}
\end{figure}
Figure~\ref{fig:error_time} shows that NeMDO$_{p=2}^1$ achieves substantially lower error than the SPH kernels at comparable or lower cost, and yields a speedup of roughly $10$ times over LABFM$_{p=2}$ up to its limiting error of $4\times 10^{-3}$. NeMDO$_{p=2}^2$ offers a more balanced trade-off, with a $1.5$–$2$ times speedup relative to LABFM$_{p=2}$ and up to two orders of magnitude lower error than the SPH baselines. Increasing the number of network parameters further in NeMDO$_{p=2}^3$ reduces the error floor again, at the expense of increased forward time.

It is important to note that all NeMDO models in Figure~\ref{fig:error_time} were trained with geometric noise of magnitude $\epsilon = 1.0$, so they retain their accuracy for particle disorder up to this level. The figure reports results for $\epsilon = 0.5$, which is approximately the highest noise level at which the SPH kernels still exhibit convergence, while LABFM$_{p=2}$ (and other consistent methods) maintain the same convergence for particle disorder from $\epsilon = 0-1.0$ for second-order approximations~\citep{King2020}. Additional experiments (Section~\ref{app:noise_sensitivity}) indicate that training with milder noise makes the learning problem simpler, yielding lower errors for the same architecture. In this sense, the cost–accuracy curves in Figure~\ref{fig:error_time} provide a conservative, worst-case characterisation of NeMDO’s performance. Taken together, these results demonstrate a clear accuracy–computational cost trade-off enabled by NeMDO, which is not readily accessible to traditional mesh-free discretisation approaches.

\subsection{Application to Partial Differential Equations} 
\label{sec:tgv}
To assess the effectiveness of the learned operators in practical settings, we use them in mesh-free simulations of the weakly compressible Navier--Stokes equations in an Eulerian unstructured node distribution. While SPH is generally implemented in Lagrangian schemes, particle clustering and instabilities can significantly affect the quality of the simulation~\citep{Price2012, Sun2018, Lyu2022}. Our objective in this work is to assess the quality of the differential operator; thus, we restrict our investigation to Eulerian simulations, and consider the two-dimensional Taylor--Green vortex. 

Weakly compressible SPH is known to generate pressure oscillations; thus, to improve numerical stability we dealiase the solution with a high-order filter at each time-step~\citep{Jameson1981}. Furthermore, high-order collocated methods are known to frequently lead to spurious oscillations at small scales~\citep{Lamballais2011}; thus, we also use a high-order filter in LABFM, and for our framework we train a hyperviscous operator $\text{GNN}^\text{hyp}_{p=4}$.

\textbf{Governing Equations: }We solve the two-dimensional compressible Navier--Stokes equations in conservative form
\begin{subequations}
    \begin{equation}
        \frac{\partial\rho}{\partial t} + \frac{\partial \rho u_k}{\partial x_k} = 0 
    \end{equation}
    \begin{equation}
        \frac{\partial \rho u_i}{\partial t} + \frac{\partial \rho u_i u_k}{\partial x_k} = -  \frac{1}{\text{Ma}^2} \frac{\partial \rho}{\partial x_i} + \frac{1}{\text{Re}}\frac{\partial^{2}u_{i}}{\partial{x}_{k}\partial{x}_{k}}
    \end{equation}
\end{subequations}
where $\rho$ is the density, $u_i$ is the $i$-th component velocity, and Re is the Reynolds number, and Ma is the Mach number. The system is closed with a barotropic equation of state absorbed into the momentum equation. The analytical solution of the TGV for incompressible flow is given by:
\begin{subequations}
\label{eq:tgv_sol}
\begin{equation}
u(x,y,t) = -\exp\!\left(-\frac{8\pi^2}{\mathrm{Re}}\,t^*\right)\cos(2\pi x)\sin(2\pi y)
\end{equation}
\begin{equation}
v(x,y,t) = \phantom{-}\exp\!\left(-\frac{8\pi^2}{\mathrm{Re}}\,t^*\right)\sin(2\pi x)\cos(2\pi y),
\end{equation}
\end{subequations}
on the spatial domain $(x,y) \in [0,1]^2$ equipped with periodic boundary conditions. In the equation above, $u$ and $v$ denote the velocity components in the $x$- and $y$-directions, respectively, which we use as a reference. In the experiments carried in this section, we used a Reynolds number of 100 and Mach number of 0.1. In the weakly compressible regime, the numerical discretisation error dominates the deviation introduced by compressibility effects; therefore, the incompressible solution can be used as a reference~\citep{King2022}.

\begin{figure}[!tbp]
\begin{center}
\makebox[0.2\linewidth][c]{$t^*=0$}
\makebox[0.2\linewidth][c]{$t^*=3$}
\makebox[0.2\linewidth][c]{$t^*=6$}
\makebox[0.2\linewidth][c]{$t^*=10$}\par

\makebox[0pt][r]{\rotatebox{90}{\small NeMDO$_{p=2}$, $s=1/70$}\hspace{0.6em}}%
\begin{subfigure}{0.2\linewidth}
  \centering
  \includegraphics[width=\linewidth]{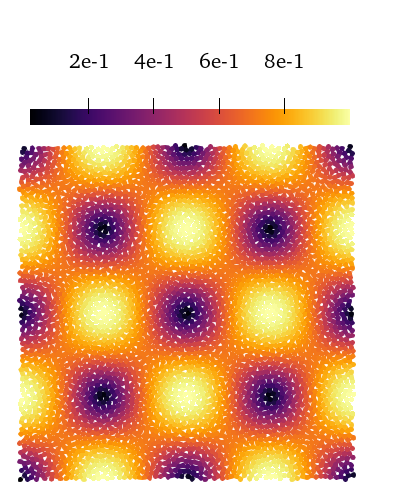}
\end{subfigure}
\begin{subfigure}{0.2\linewidth}
  \centering
  \includegraphics[width=\linewidth]{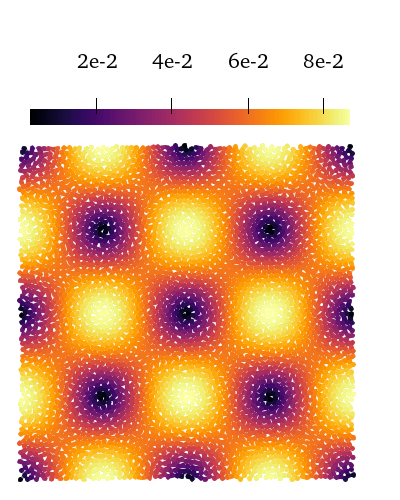}
\end{subfigure}
\begin{subfigure}{0.2\linewidth}
  \centering
  \includegraphics[width=\linewidth]{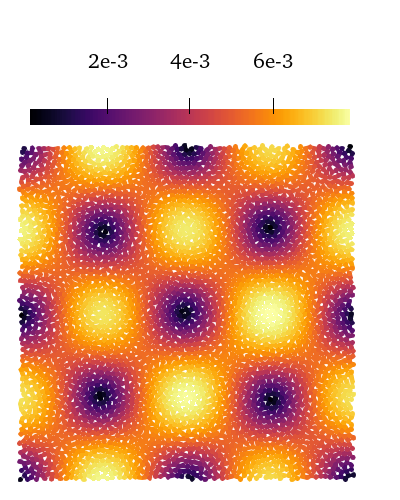}
\end{subfigure}
\begin{subfigure}{0.2\linewidth}
  \centering
  \includegraphics[width=\linewidth]{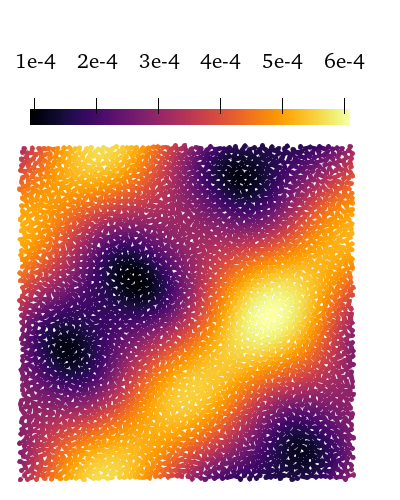}
\end{subfigure}

\makebox[0pt][r]{\rotatebox{90}{\small Wendland C2, $s=1/70$}\hspace{0.6em}}%
\begin{subfigure}{0.2\linewidth}
  \centering
  \includegraphics[width=\linewidth]{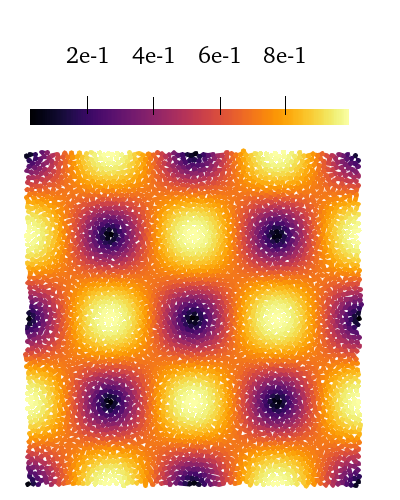}
\end{subfigure}
\begin{subfigure}{0.2\linewidth}
  \centering
  \includegraphics[width=\linewidth]{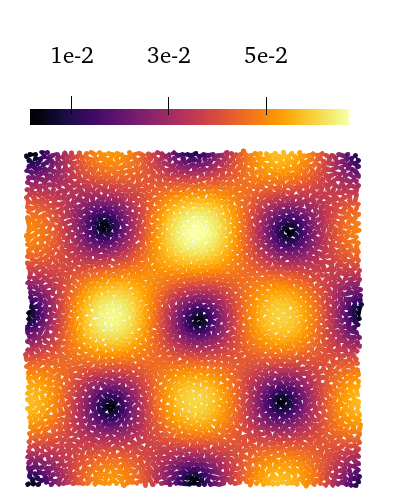}
\end{subfigure}
\begin{subfigure}{0.2\linewidth}
  \centering
  \includegraphics[width=\linewidth]{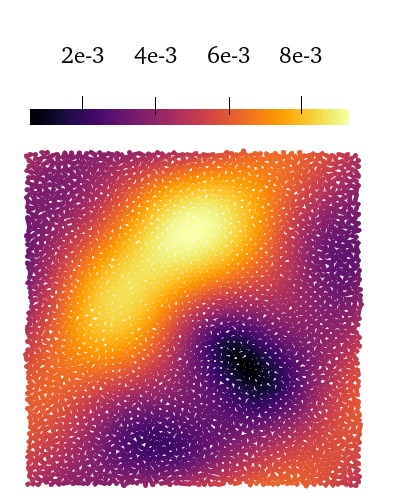}
\end{subfigure}
\begin{subfigure}{0.2\linewidth}
  \centering
  \includegraphics[width=\linewidth]{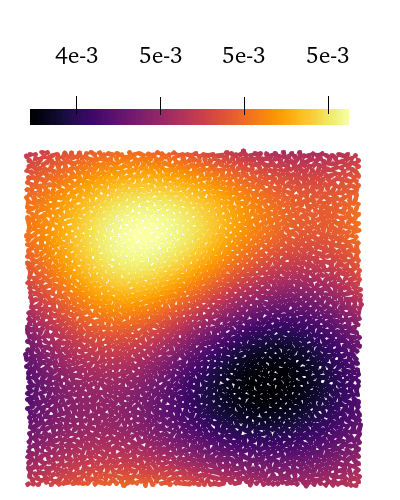}
\end{subfigure}

\makebox[0pt][r]{\rotatebox{90}{\small NeMDO$_{p=2}$, $s=1/140$}\hspace{0.6em}}%
\begin{subfigure}{0.2\linewidth}
  \centering
  \includegraphics[width=\linewidth]{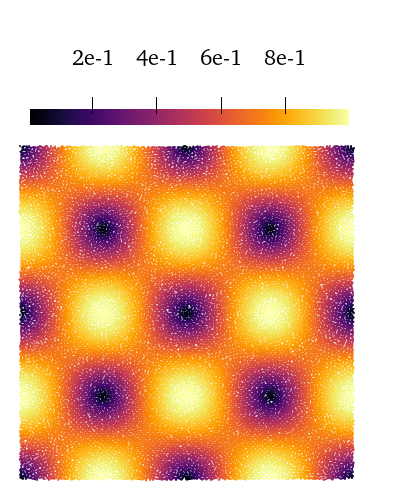}
\end{subfigure}
\begin{subfigure}{0.2\linewidth}
  \centering
  \includegraphics[width=\linewidth]{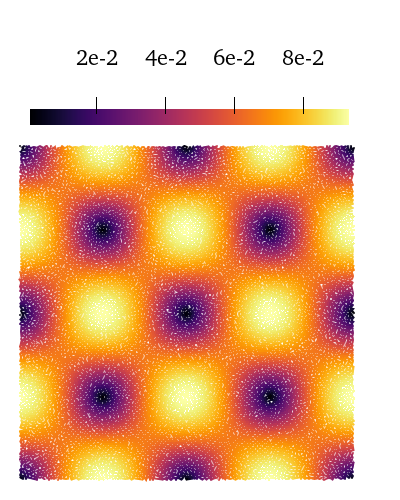}
\end{subfigure}
\begin{subfigure}{0.2\linewidth}
  \centering
  \includegraphics[width=\linewidth]{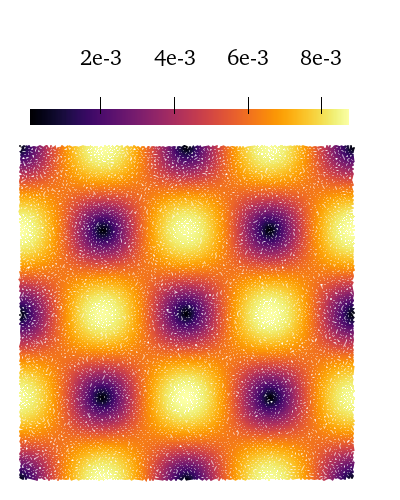}
\end{subfigure}
\begin{subfigure}{0.2\linewidth}
  \centering
  \includegraphics[width=\linewidth]{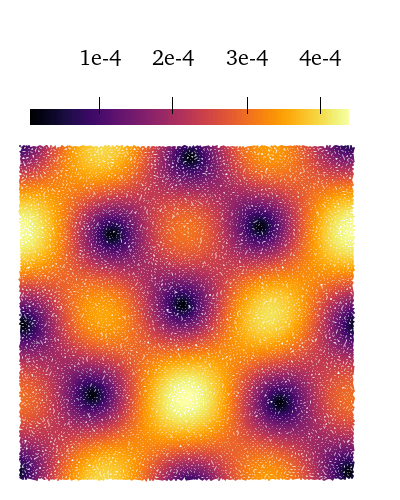}
\end{subfigure}

\makebox[0pt][r]{\rotatebox{90}{\small Wendland C2, $s=1/140$}\hspace{0.6em}}%
\begin{subfigure}{0.2\linewidth}
  \centering
  \includegraphics[width=\linewidth]{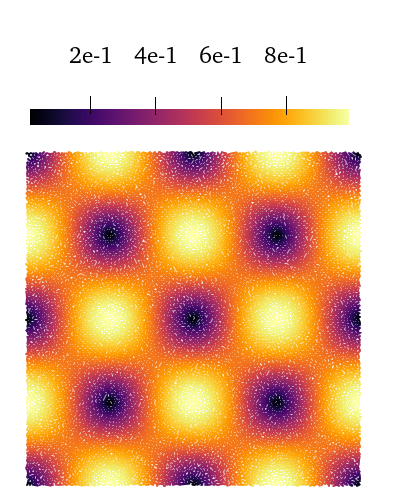}
\end{subfigure}
\begin{subfigure}{0.2\linewidth}
  \centering
  \includegraphics[width=\linewidth]{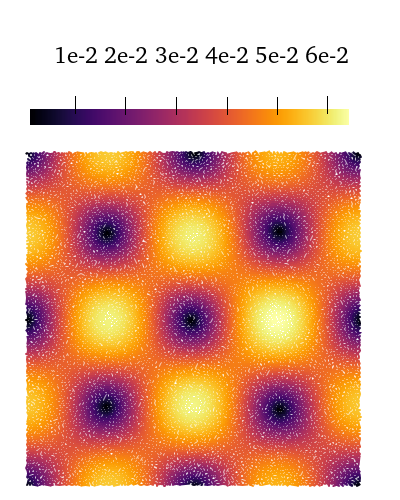}
\end{subfigure}
\begin{subfigure}{0.2\linewidth}
  \centering
  \includegraphics[width=\linewidth]{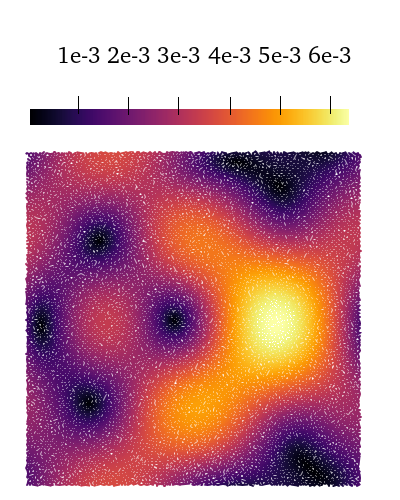}
\end{subfigure}
\begin{subfigure}{0.2\linewidth}
  \centering
  \includegraphics[width=\linewidth]{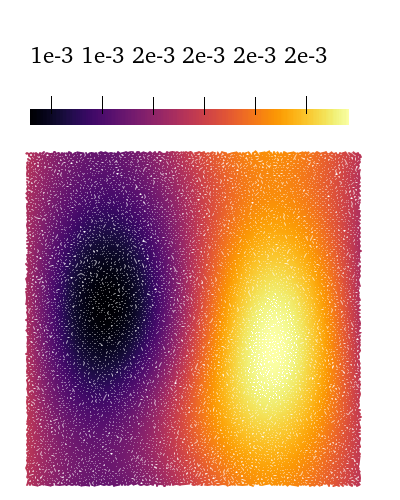}
\end{subfigure}

\makebox[0pt][r]{\rotatebox{90}{\small LABFM$_{p=4}$, $s=1/70$}\hspace{0.6em}}%
\begin{subfigure}{0.2\linewidth}
  \centering
  \includegraphics[width=\linewidth]{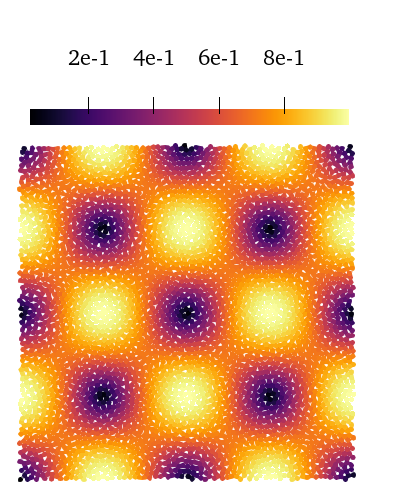}
\end{subfigure}
\begin{subfigure}{0.2\linewidth}
  \centering
  \includegraphics[width=\linewidth]{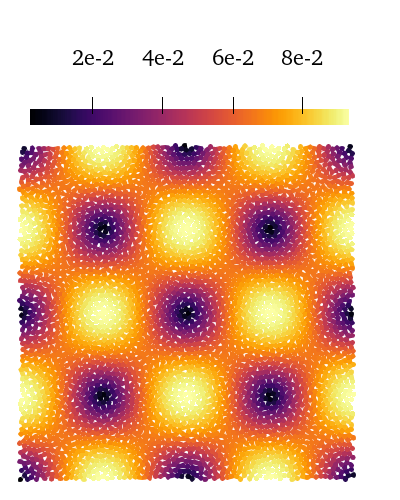}
\end{subfigure}
\begin{subfigure}{0.2\linewidth}
  \centering
  \includegraphics[width=\linewidth]{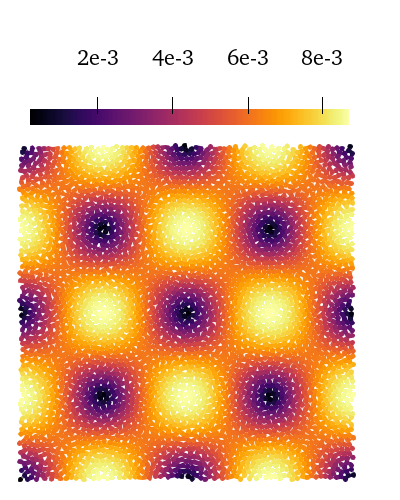}
\end{subfigure}
\begin{subfigure}{0.2\linewidth}
  \centering
  \includegraphics[width=\linewidth]{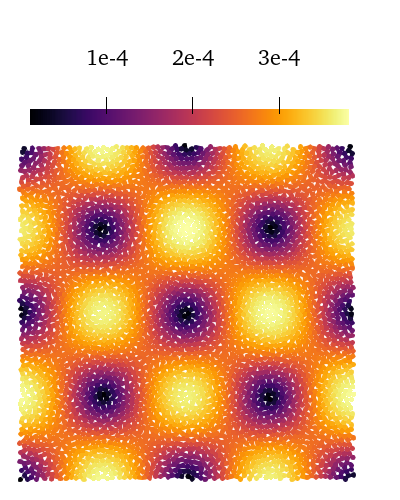}
\end{subfigure}

\end{center}
\caption{Velocity magnitude for the Taylor--Green vortex with different operators at different resolutions. Colour axis are rescaled at different time scales for better visualisation of the flow structures.}
\label{fig:tgv_vel}
\end{figure}

Figure~\ref{fig:tgv_vel} shows snapshots of the velocity magnitude of the TGV at different time units for the SPH operator with Wendland C2 kernel, NeMDO$_{p=2}$, and LABFM$_{p=4}$. We simulate the TGV with the SPH operator and NeMDO for coarse and fine resolutions to test the convergence, while LABFM$_{p=4}$ as a representative high-order mesh-free discretisation.

Qualitatively, NeMDO outperforms SPH for both coarse and fine resolutions. While SPH at $s=1/70$ shows asymmetry in the flow at 3 time units, our operator at the same resolution only shows asymmetry after 10 time units. At higher resolutions, our operator shows a significant improvement in capturing the large flow structure compared to the lower resolution, indicating convergence. SPH does not show significant improvement when increasing resolution. LABFM outperforms both methods, which is expected due to the fourth-order accuracy enforced. These results agree with the limiting errors observed earlier in the convergence study, showing SPH low limiting error, while NeMDO shows a clear convergent behavior and significant improvement over SPH.

To perform a quantitative assessment of the results, we compute the error between the analytical solution and the discrete operators with a relative root mean squared error (relative $L_2$ error) based on the velocity magnitude. 


\begin{figure}[ht]
  \begin{center}
  \includegraphics[width=0.6\linewidth]{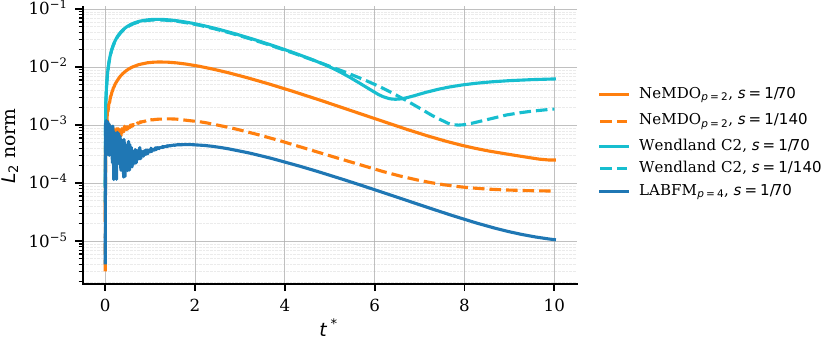}
  \caption{Relative root mean squared error of different operators with respect to the analytical solution (Equation~\eqref{eq:tgv_sol}) for velocity. $s$ indicates the average particle spacing, and $t^*$ indicates characteristic time units of the flow.
    }
    \label{fig:tgv_analytic}
  \end{center}
\end{figure}

Figure~\ref{fig:tgv_analytic} reports the relative root-mean-square error of the TGV. These results confirm the observations conducted in the qualitative analysis, where both SPH and NeMDO exhibit convergent behaviour with increasing resolution; however, the SPH error saturates at order $10^{-3}$, whereas NeMDO achieves errors approximately one order of magnitude smaller across the tested resolutions. Indeed, the observed performance of NeMDO is consistent with the operator-level accuracy and stability diagnostics reported earlier, indicating that the learnt operators can be directly embedded within a compressible flow solver and used as a standalone spatial discretisation without problem-specific tuning. In contrast, SPH requires either high-order filtering (provided by a high-order consistent method) or kernel correction to obtain stable solutions. Because NeMDO is trained without access to PDE solution data, the learnt operators can be deployed in a zero-shot manner at the differential operator level in mesh-free simulations, enabling direct reuse within PDE solvers without retraining or further changes in the code.

\section{Parametric Study}
\label{app:ablation} 
To assess the robustness and consistency of the proposed operator construction, we conduct a parametric study examining the influence of: (i) stencil size, (ii) Taylor truncation, and (iii) particle disorder.
\label{app:ablation}
\label{sec:ablation}
\subsection{Stencil Size}
We first study how the accuracy of the learned operators depends on the stencil size $|\mathcal{N}_i|$. Table~\ref{tab:stencil_size} reports the average moment error and standard deviation for $x$-derivative operators as $|\mathcal{N}_i|$ is varied for disordered node distribution, while keeping the network architecture and all other training parameters fixed.

\begin{table}[t]
\caption{NeMDO moment residuals for $x$-derivative operators with varying stencil size $|\mathcal{N}_i|$. Mean absolute error (MAE) and standard deviation are reported for each monomial moment, averaged over independently perturbed neighbourhood realisations. Residuals quantify the extent to which the learned discrete operators satisfy the imposed Taylor consistency constraints (smaller values are better).}
\label{tab:stencil_size}
\begin{center}
\setlength{\tabcolsep}{8pt} 
\renewcommand{\arraystretch}{1.1} 
\begin{tabular}{cccccccc}
\multicolumn{1}{c}{\bf $|\mathcal{N}_i|$} &
\multicolumn{1}{c}{\bf Operator} &
\multicolumn{1}{c}{\bf Metric} &
\multicolumn{1}{c}{\bf $x$} &
\multicolumn{1}{c}{\bf $y$} &
\multicolumn{1}{c}{\bf $x^2/2$} &
\multicolumn{1}{c}{\bf $xy$} &
\multicolumn{1}{c}{\bf $y^2/2$}
\\ \hline \\

\multirow{2}{*}{10} & \multirow{2}{*}{$\text{NeMDO}_{p=2}$}& MAE  & $6.44 \times 10^{-3}$ & $6.58 \times 10^{-3}$ & $9.56 \times 10^{-3}$ & $9.97 \times 10^{-3}$ & $9.95 \times 10^{-3}$ \\
                      && St.d. & $9.89 \times 10^{-3}$ & $1.05 \times 10^{-2}$ & $1.76 \times 10^{-2}$ & $1.70 \times 10^{-2}$ & $1.63 \times 10^{-2}$ \\
\hline
\multirow{2}{*}{25} & \multirow{2}{*}{$\text{NeMDO}_{p=2}$}& MAE  & $7.84 \times 10^{-4}$ & $8.12 \times 10^{-4}$ & $5.23 \times 10^{-4}$ & $6.47 \times 10^{-4}$ & $6.29 \times 10^{-4}$ \\
                           && St.d. & $1.06 \times 10^{-3}$ & $1.10 \times 10^{-3}$ & $7.21 \times 10^{-4}$ & $9.13 \times 10^{-4}$ & $8.91 \times 10^{-4}$ \\
                           \hline
\multirow{2}{*}{50} & \multirow{2}{*}{$\text{NeMDO}_{p=2}$}& MAE  & $4.43 \times 10^{-4}$ & $3.95 \times 10^{-4}$ & $4.00 \times 10^{-4}$ & $4.15 \times 10^{-4}$ & $4.14 \times 10^{-4}$ \\
                           && St.d. & $5.68 \times 10^{-4}$ & $5.18 \times 10^{-4}$ & $5.25 \times 10^{-4}$ & $5.76 \times 10^{-4}$ & $5.60 \times 10^{-4}$ \\
                           \hline
\multirow{2}{*}{100} & \multirow{2}{*}{$\text{NeMDO}_{p=2}$}& MAE  & $1.25 \times 10^{-3}$ & $1.24 \times 10^{-3}$ & $1.63 \times 10^{-3}$ & $1.79 \times 10^{-3}$ & $1.59 \times 10^{-3}$ \\
                           && St.d. & $1.61 \times 10^{-3}$ & $1.56 \times 10^{-3}$ & $2.15 \times 10^{-3}$ & $2.29 \times 10^{-3}$ & $2.03 \times 10^{-3}$ \\

\end{tabular}
\end{center}
\end{table}

\begin{figure}[t]
\begin{center}
  \begin{subfigure}[t]{0.24\linewidth}
    \begin{center}
      \includegraphics[width=\linewidth]{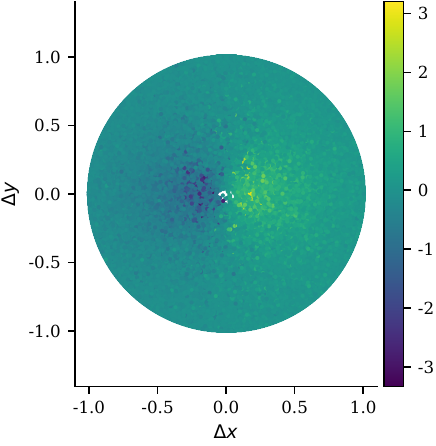}
    \end{center}
    \caption{$|\mathcal{N}_i|=10$}
  \end{subfigure}\hfill
  \begin{subfigure}[t]{0.24\linewidth}
    \begin{center}
      \includegraphics[width=\linewidth]{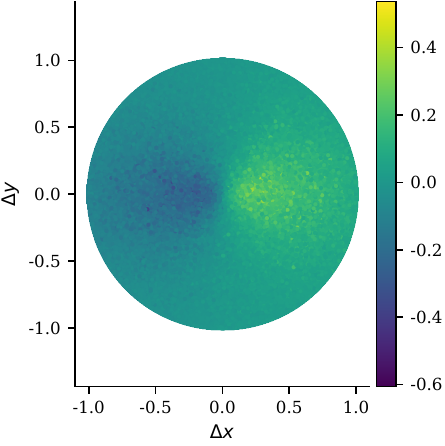}
    \end{center}
    \caption{$|\mathcal{N}_i|=25$}
  \end{subfigure}\hfill
  \begin{subfigure}[t]{0.24\linewidth}
    \begin{center}
      \includegraphics[width=\linewidth]{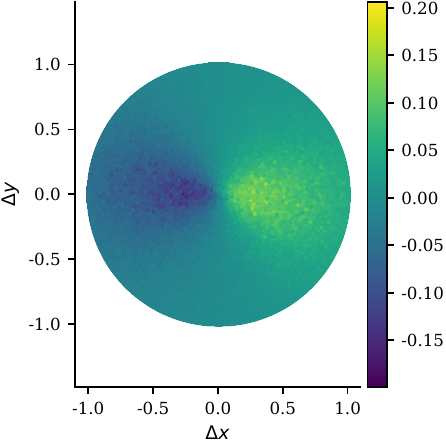}
    \end{center}
    \caption{$|\mathcal{N}_i|=50$}
  \end{subfigure}\hfill
  \begin{subfigure}[t]{0.24\linewidth}
    \begin{center}
      \includegraphics[width=\linewidth]{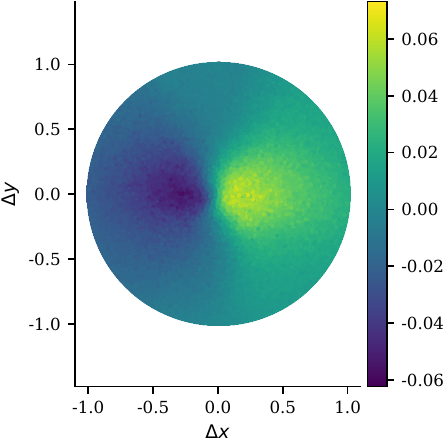}
    \end{center}
    \caption{$|\mathcal{N}_i|=100$}
  \end{subfigure}
\end{center}
\caption{Parametric results for varying stencil size $n$ for node disorder of $\epsilon=1.0$. Colours indicate value of weights. To generate the plots, multiple stencils with normalised relative positions have their weights predicted and plotted on top of each other.}
\label{fig:appendix_ablation_n_p_grid}
\end{figure}

Table~\ref{tab:stencil_size} shows that increasing $|\mathcal{N}_i|$ from $10$ up to $50$ neighbours reduces the average moment error, this reflects the improved conditioning of the local Taylor system when more neighbours are included (indicated by the smaller weights in Figure~\ref{fig:appendix_ablation_n_p_grid}). However, further increasing the stencil size to $|\mathcal{N}_i|=100$ does not lead to additional gains. This suggests the presence of an optimal stencil size, beyond which particle noise, network capacity, and training dataset size, rather than the number of neighbours, become the dominant limiting factors. Consistent with the convergence results in Section~\ref{sec:convergence}, larger residuals (reported in Table~\ref{tab:moments-models}) translate into lower accuracy. Consequently, models exhibiting larger moment residuals demonstrate a higher error floor.

Figure~\ref{fig:appendix_ablation_n_p_grid} shows the plot of multiple stencil configurations with the particles coloured by weight on top of each other. Stencils with a smaller neighbour count exhibit high-frequency spatial fluctuations and a lack of discernible topological structure. This noisy distribution implies that the local consistency conditions are satisfied through highly non-symmetric weight assignments. As neighbour count increases, the stencils undergo a clear structural refinement, transitioning from a noisy distribution of weights toward an anti-symmetric stencil structure. Furthermore, the magnitude of the weights decreases with increasing neighbour count, reflecting a transition from an ill-conditioned local approximation to a more regularised and distributed representation. 

\begin{figure}[h]
  \begin{center}
  \includegraphics[width=0.45\linewidth]{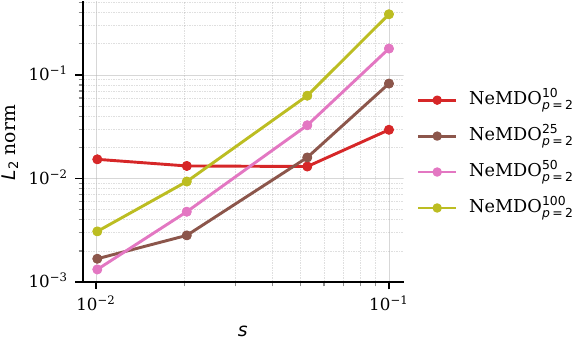}
  \caption{Convergence of the discrete $x$-derivative operator on a smooth test function with $\epsilon = 1.0$. Relative $L_2$ norm versus particle spacing $s$ for the learned NeMDO operator with varying stencil size. The superscript in the operator indicates the number of particles within the computational stencil.}
    \label{fig:stencil_conv}
  \end{center}
\end{figure}
Figure~\ref{fig:stencil_conv} illustrates the convergence behavior of the NeMDO $x$-derivative operator as a function of the neighborhood size $|\mathcal{N}_i|$. At lower resolutions ($s=10^{-1}$), stencils with a smaller number of neighbors yield superior $L_2$ error norms; however, as the resolution increases, larger stencils demonstrate a more favorable asymptotic error floor. This observed stagnation in accuracy is consistent with the moment analysis presented in Table~\ref{tab:stencil_size}. For example, the first-order moment sum for NeMDO$_{p=2}^{10}$ is approximately $1.3 \times 10^{-2}$, which directly corresponds to the resolution at which the operator reaches its formal truncation error limit.
 
\begin{figure}[h]
  \begin{center}
    \begin{subfigure}{0.44\linewidth}
      \centering
      \includegraphics[width=\linewidth]{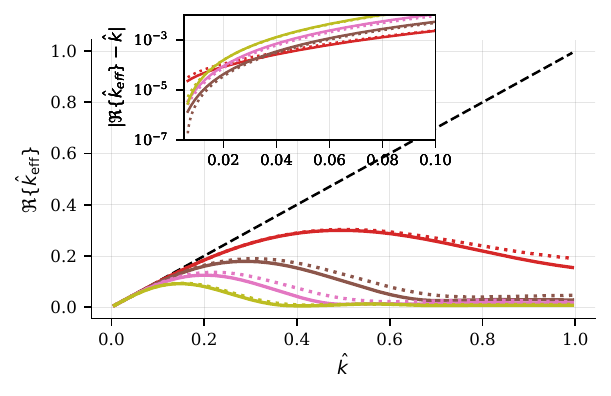}
    \end{subfigure}
    \begin{subfigure}{0.54\linewidth}
      \centering
      \includegraphics[width=\linewidth]{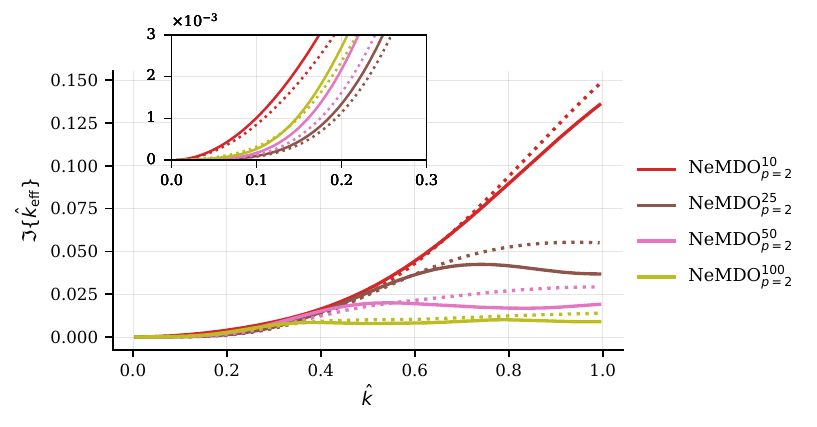}
    \end{subfigure}
  \end{center}
  \caption{Resolving power of the gradient operator for varying number of particles in computational stencil. The \textit{left} plot shows the real part of the modal response, and the \textit{right} plot shows the imaginary part on disordered particle distributions with noise level $\epsilon = 1.0$. The horizontal axis shows the true wavenumber $k$ normalised by the Nyquist wavenumber $k_{\mathrm{Ny}}$, while the vertical axis shows the corresponding effective wavenumber $k_{\mathrm{eff}}/k_{\mathrm{Ny}}$. The dashed black line indicates the ideal response, corresponding to a spectral method. The inset on the \textit{left} plot shows the absolute difference between the different operators and the ideal spectral response. The inset on the \textit{right} plot shows zoomed in graph of $\Im\{\hat{k}_{eff}\}$ for range $0\leq\hat{k}\leq 0.3$. The full and dotted lines indicate $k_y/k_x=0$ and $k_y/k_x=1$, respectively.}
  \label{fig:res_stencil}
\end{figure}
Figure~\ref{fig:res_stencil} illustrates the resolving power of NeMDO across varying neighborhood sizes. At higher wavenumbers, reducing the stencil size significantly improves the real component of the modal response. This spectral behavior directly correlates with the lower $L_2$ error norms observed at coarser resolutions in Figure~\ref{fig:stencil_conv}. However, at low wavenumbers (Figure~\ref{fig:res_stencil}, \textit{left} inset), the NeMDO$_{p=2}^{10}$ operator exhibits a larger absolute error compared to configurations withlarger supports. This deterioration in the low-frequency modal response is a consequence of larger moment residuals, which impose a stricter truncation error limit. Conversely, increasing the stencil size leads to a reduction in dissipation error. These results suggest a fundamental trade-off between improved modal response for high-wavenumber under-constrained stencils and the improved consistency afforded by larger neighborhood sizes.

\subsection{Taylor Truncation}
We now vary the Taylor truncation by increasing the number of moments included in $\mathbf{M}^D$. In traditional high-order mesh-free numerical methods, this enforces polynomial consistency to a higher degree, which leads to more accurate solutions at equivalent resolution (i.e. improved resolving power). However, the local systems that must be solved become more poorly conditioned for higher orders of approximation; thus, these usually require larger computational stencils, and are more prone to instabilities~\citep{shankar2014radialbasisfunctionrbffinite, King2020}.

\begin{figure}[t]
  \begin{center}
  \includegraphics[width=0.3\linewidth]{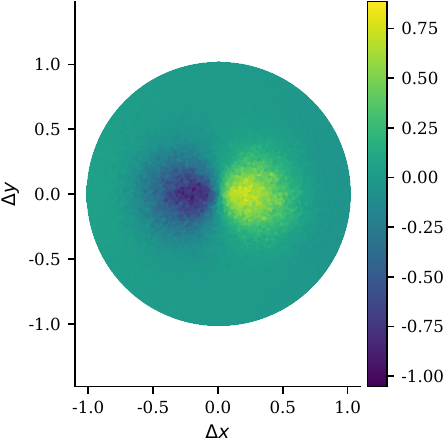}
  \caption{Normalised third-order NeMDO for the $x-$derivative, colours indicate weight value. Model was trained with $|\mathcal{N}_i|=50$ and particle disturbance of $\epsilon=1.0$ (and inferred with the same noise).}
    \label{fig:p3_stencil}
  \end{center}
\end{figure}

Figure~\ref{fig:p3_stencil} illustrates the stencil morphology for the NeMDO $x$-derivative operator trained with third-order Taylor monomials and $|\mathcal{N}_i|=50$. Consistent with the second-order approximation (Figure~\ref{fig:appendix_ablation_n_p_grid}c), the third-order operator maintains a defined anti-symmetric structure about the origin. However, the higher-order consistency requirement leads to increased weight magnitudes compared to the second-order counterpart. Furthermore, these larger weights exhibit a higher spatial concentration closer to the origin. This suggests that as the polynomial order increases, the operator prioritizes the nearest neighbors to satisfy the more stringent higher-order moments.

Table~\ref{tab:moments-p3} shows NeMDO average moments up to third-order Taylor monomials averaged over multiple stencil configurations unseen during training. All moments are satisfied with similar accuracy approximately at order $10^{-3}$. Compared to the second-order approximation with the same stencil size (in Table~\ref{tab:stencil_size}), the moments are about one order of magnitude larger. By increasing the terms present in the Taylor truncation, the loss function becomes more complex, requiring the optimizer to balance a larger set of competing consistency conditions within the same local support. This increased requirement for higher-order polynomial accuracy causes degradation in moment residuals.

\begin{table}[t]
\caption{NeMDO moment residuals for $x$-derivative operators with Taylor monomials up to third order, $\text{NeMDO}_{p=3}^x$. Mean absolute error (MAE) and standard deviation are reported for each monomial moment, averaged over independently perturbed neighbourhood realisations. Residuals quantify the extent to which the learned discrete operators satisfy the imposed Taylor consistency constraints (smaller values are better).}
\label{tab:moments-p3}
\begin{center}
\setlength{\tabcolsep}{2pt} 
\renewcommand{\arraystretch}{1.1} 
\begin{tabular}{cccccccccccc}
\multicolumn{1}{c}{\bf Metric} &
\multicolumn{1}{c}{\bf $x$} &
\multicolumn{1}{c}{\bf $y$} &
\multicolumn{1}{c}{\bf $x^2/2$} &
\multicolumn{1}{c}{\bf $xy$} &
\multicolumn{1}{c}{\bf $y^2/2$} &
\multicolumn{1}{c}{\bf $x^3/6$} & 
\multicolumn{1}{c}{\bf $x^2y/2$} & 
\multicolumn{1}{c}{\bf $xy^2/2$} & 
\multicolumn{1}{c}{\bf $y^3/6$} & 
\\ \hline \\

 MAE   & $2.08 \times 10^{-3}$ & $2.07 \times 10^{-3}$ & $1.71 \times 10^{-3}$ & $1.92 \times 10^{-3}$ & $1.85 \times 10^{-3}$ & $2.42 \times 10^{-3}$ & $2.43 \times 10^{-3}$ & $2.39 \times 10^{-3}$ & $2.86 \times 10^{-3}$\\
                      St.d. & $2.70 \times 10^{-3}$ & $2.73 \times 10^{-3}$ & $2.24 \times 10^{-3}$ & $2.52 \times 10^{-3}$ & $2.40 \times 10^{-3}$ & $3.29 \times 10^{-3}$ & $3.21 \times 10^{-3}$ & $3.21 \times 10^{-3}$ & $ 3.61 \times 10^{-3}$\\

\end{tabular}
\end{center}
\end{table}

\begin{figure}[ht]
  \begin{center}
  \includegraphics[width=0.45\linewidth]{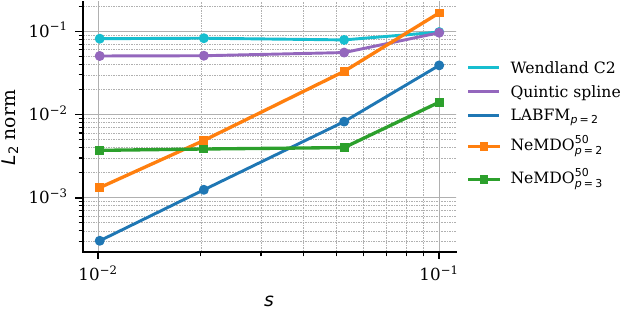}
  \caption{Convergence of the discrete $x$-derivative operator on a smooth test function with $\epsilon = 0.5$. Relative $L_2$ norm versus particle spacing $s$ for the learned NeMDO operator with varying Taylor truncation. The superscript in the NeMDO operators indicate the stencil size $|\mathcal{N}_i|$.}
    \label{fig:p3_conv}
  \end{center}
\end{figure}

Figure~\ref{fig:p3_conv} illustrates the convergence behavior of the third-order NeMDO $x$-derivative operator. At lower resolutions, NeMDO$_{p=3}^{50}$ outperforms all other operators, demonstrating the benefits of increasing the Taylor truncation.
However, as the resolution increases, the model trained with more stencil moments quickly reaches its limiting error, indicated by the sum of the first-order moments in Table~\ref{tab:moments-p3}. 

\begin{figure}[t]
  \begin{center}
    \begin{subfigure}{0.44\linewidth}
      \centering
      \includegraphics[width=\linewidth]{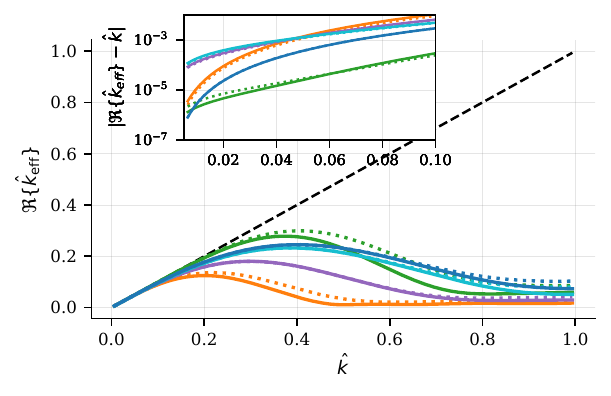}
    \end{subfigure}
    \begin{subfigure}{0.54\linewidth}
      \centering
      \includegraphics[width=\linewidth]{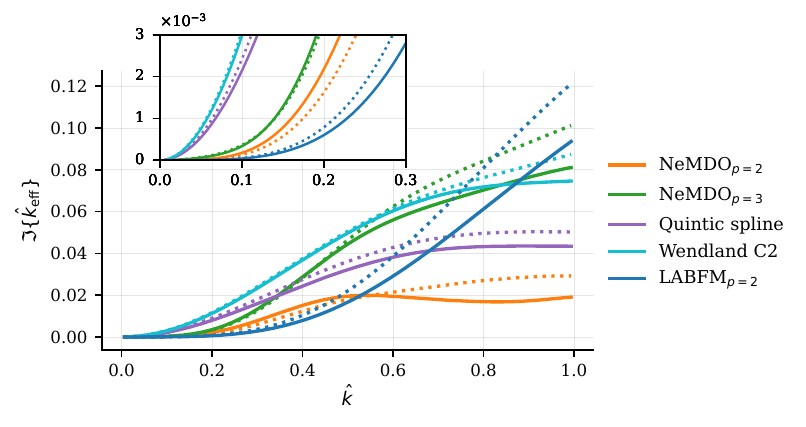}
    \end{subfigure}
  \end{center}
  \caption{Resolving power of the gradient operator for increasing Taylor truncation included during training. The \textit{left} plot shows the real part of the modal response, and the \textit{right} plot shows the imaginary part on disordered particle distributions with noise level $\epsilon = 1.0$. The horizontal axis shows the true wavenumber $k$ normalised by the Nyquist wavenumber $k_{\mathrm{Ny}}$, while the vertical axis shows the corresponding effective wavenumber $k_{\mathrm{eff}}/k_{\mathrm{Ny}}$. The dashed black line indicates the ideal response, corresponding to a spectral method. The inset on the \textit{left} plot shows the absolute difference between the different operators and the ideal spectral response. The inset on the \textit{right} plot shows zoomed in graph of $\Im\{\hat{k}_{eff}\}$ for range $0\leq\hat{k}\leq 0.3$. The full and dotted lines indicate $k_y/k_x=0$ and $k_y/k_x=1$, respectively.}
  \label{fig:ablation_trunc}
\end{figure}

Figure~\ref{fig:ablation_trunc} (\textit{left}) illustrates the resolving power as a function of the Taylor truncation order. Consistent with classical high-order numerical schemes, extending the truncated expansion improves the spectral fidelity of the real component of the modal response. This enhancement is further evidenced by the convergence results in Figure~\ref{fig:p3_conv}, where the $p=3$ model significantly outperforms the lower-order operators at coarse resolutions. However, for wavenumbers $\hat{k} \geq 0.5$, the real component of the NeMDO${p=3}$ response begins to attenuate, falling below the performance of both LABFM${p=2}$ and the Wendland C2 kernel. At low wavenumbers ($\hat{k} \leq 0.04$), the inset of Figure~\ref{fig:ablation_trunc} (\textit{left}) reveals a distinct divergence in error scaling. For the NeMDO$_{p=2}$ and LABFM$_{p=2}$ operators, the error magnitude exhibits a steep positive gradient for $\hat{k} \in [0, 0.04]$, which subsequently transitions to a lower growth rate as $\hat{k}$ increases. This initial steepness reflects the operators' ability to resolve increasingly smooth modes with high precision. In contrast, the error profile for NeMDO$_{p=3}$ remains relatively invariant across this low-frequency regime. This plateau indicates that the $p=3$ operator has reached a truncation error floor, where the error is no longer dominated by the wavenumber but by the constant moment residuals reported in Table~\ref{tab:moments-p3}. A similar spectral stagnation is observed for the SPH kernels, which exhibit a horizontal error profile as $\hat{k} \to 0$, indicating a lack of formal consistency at the lowest frequencies.

\subsection{Particle Disorder}
\label{app:noise_sensitivity}
This section investigates the sensitivity of the NeMDO operators to varying degrees of geometric disorder, $\epsilon$, during both the training and inference phases. To evaluate the generalization capabilities of the learned weights, we trained separate NeMDO models across a spectrum of particle perturbations, specifically $\epsilon \in \{0.1, 0.5, 1.0\}$. We then performed a cross-evaluation to determine the impact of training-set stochasticity on the resulting operator consistency. Table~\ref{tab:noisy_moments} summarizes the moment residuals for each configuration, providing a quantitative measure of how well each model maintains polynomial consistency when subjected to different noise regimes.

\begin{table}[t]
\caption{NeMDO moment residuals for $x$-derivative operators with varying particle disorder $\epsilon$. Mean absolute error (MAE) and standard deviation are reported for each monomial moment, averaged over independently perturbed neighbourhood realisations. Residuals quantify the extent to which the learned discrete operators satisfy the imposed Taylor consistency constraints (smaller values are better). Trained $\epsilon$ denotes to the particle disorder during training, and inferred $\epsilon$ denotes the particle disorder during inference.}
\label{tab:noisy_moments}
\begin{center}
\setlength{\tabcolsep}{5pt} 
\renewcommand{\arraystretch}{1.1} 
\begin{tabular}{ccccccccc}
\multicolumn{1}{c}{\bf Trained $\epsilon$} &
\multicolumn{1}{c}{\bf Operator} &
\multicolumn{1}{c}{\bf Inferred $\epsilon$} &
\multicolumn{1}{c}{\bf Metric} &
\multicolumn{1}{c}{\bf $x$} &
\multicolumn{1}{c}{\bf $y$} &
\multicolumn{1}{c}{\bf $x^2/2$} &
\multicolumn{1}{c}{\bf $xy$} &
\multicolumn{1}{c}{\bf $y^2/2$}
\\ \hline \\

\multirow{6}{*}{1.0} & \multirow{6}{*}{$\text{NeMDO}_{p=2}$} & \multirow{2}{*}{1.0}                             & MAE  & $7.88 \times 10^{-4}$ & $7.73 \times 10^{-4}$ & $8.51 \times 10^{-4}$ & $8.86 \times 10^{-4}$ & $8.91 \times 10^{-4}$ \\
                      &&& St.d. & $1.07 \times 10^{-3}$ & $9.89 \times 10^{-4}$ & $1.21 \times 10^{-3}$ & $1.24 \times 10^{-3}$ & $1.22 \times 10^{-3}$ \\
                      \cline{3-9}
                      && \multirow{2}{*}{0.5} & MAE  & $4.38 \times 10^{-4}$ & $4.88 \times 10^{-4}$ & $4.23 \times 10^{-4}$ & $3.93 \times 10^{-4}$ & $4.81 \times 10^{-4}$ 
                      \\&&& St.d. & $4.74 \times 10^{-4}$ & $4.78 \times 10^{-4}$ & $5.26 \times 10^{-4}$ & $5.04 \times 10^{-4}$ & $5.79 \times 10^{-4}$ \\
                      \cline{3-9}
                    && \multirow{2}{*}{0.1} & MAE  & $3.60 \times 10^{-4}$ & $5.83 \times 10^{-4}$ & $2.57 \times 10^{-4}$ & $3.56 \times 10^{-4}$ & $3.48 \times 10^{-4}$ 
                      \\&&& St.d. & $1.03 \times 10^{-4}$ & $9.63 \times 10^{-5}$ & $1.76 \times 10^{-4}$ & $1.43 \times 10^{-4}$ & $1.66 \times 10^{-4}$ \\
\hline
\multirow{6}{*}{0.5} & \multirow{6}{*}{$\text{NeMDO}_{p=2}$} & \multirow{2}{*}{1.0}                             & MAE  & $2.35 \times 10^{-3}$ & $1.73 \times 10^{-3}$ & $1.64 \times 10^{-3}$ & $1.75 \times 10^{-3}$ & $1.52 \times 10^{-3}$ \\
                      &&& St.d. & $3.52 \times 10^{-3}$ & $2.47 \times 10^{-3}$ & $2.50 \times 10^{-3}$ & $2.67 \times 10^{-3}$ & $2.26 \times 10^{-3}$ \\
                      \cline{3-9}
                      && \multirow{2}{*}{0.5} & MAE  & $4.74 \times 10^{-4}$ & $4.41 \times 10^{-4}$ & $4.63 \times 10^{-4}$ & $5.13 \times 10^{-4}$ & $4.83 \times 10^{-4}$ 
                      \\&&& St.d. & $6.21 \times 10^{-4}$ & $5.88 \times 10^{-4}$ & $6.15 \times 10^{-4}$ & $6.85 \times 10^{-4}$ & $6.39 \times 10^{-4}$ \\
                      \cline{3-9}
                    && \multirow{2}{*}{0.1} & MAE  & $2.13 \times 10^{-4}$ & $5.85 \times 10^{-5}$ & $9.95 \times 10^{-5}$ & $1.08 \times 10^{-4}$ & $3.42 \times 10^{-4}$ 
                      \\&&& St.d. & $1.18 \times 10^{-4}$ & $6.95 \times 10^{-5}$ & $8.67 \times 10^{-5}$ & $1.22 \times 10^{-4}$ & $9.58 \times 10^{-5}$ \\
                           \hline
\multirow{6}{*}{0.1} & \multirow{6}{*}{$\text{NeMDO}_{p=2}$} & \multirow{2}{*}{1.0}                             & MAE  & $1.60 \times 10^{-2}$ & $1.34 \times 10^{-2}$ & $1.32 \times 10^{-2}$ & $1.39 \times 10^{-2}$ & $1.21 \times 10^{-2}$ \\
                      &&& St.d. & $2.09 \times 10^{-2}$ & $1.92 \times 10^{-2}$ & $1.89 \times 10^{-2}$ & $1.73 \times 10^{-2}$ & $1.72 \times 10^{-2}$ \\
                      \cline{3-9}
                      && \multirow{2}{*}{0.5} & MAE  & $6.11 \times 10^{-3}$ & $5.34 \times 10^{-3}$ & $6.58 \times 10^{-3}$ & $6.70 \times 10^{-3}$ & $5.74 \times 10^{-3}$ 
                      \\&&& St.d. & $8.13 \times 10^{-3}$ & $7.42 \times 10^{-3}$ & $9.46 \times 10^{-3}$ & $8.45 \times 10^{-3}$ & $8.23 \times 10^{-3}$ \\
                      \cline{3-9}
                    && \multirow{2}{*}{0.1} & MAE  & $1.09 \times 10^{-4}$ & $9.50 \times 10^{-5}$ & $1.33 \times 10^{-4}$ & $1.38 \times 10^{-4}$ & $1.43 \times 10^{-4}$ 
                      \\&&& St.d. & $1.69 \times 10^{-4}$ & $1.37 \times 10^{-4}$ & $1.97 \times 10^{-4}$ & $2.14 \times 10^{-4}$ & $1.99 \times 10^{-4}$ \\
                           \hline
                           
\multirow{6}{*}{\textbf{---}} & \multirow{6}{*}{Wendland C2} & \multirow{2}{*}{1.0}                             & MAE  & $1.01 \times 10^{-1}$ & $7.08 \times 10^{-2}$ & $4.80 \times 10^{-2}$ & $4.93 \times 10^{-2}$ & $2.55 \times 10^{-2}$ \\
                      &&& St.d. & $1.21 \times 10^{-1}$ & $8.84 \times 10^{-2}$ & $5.90 \times 10^{-2}$ & $6.16 \times 10^{-2}$ & $3.18 \times 10^{-2}$ \\
                      \cline{3-9}
                      && \multirow{2}{*}{0.5} & MAE  & $4.94 \times 10^{-2}$ & $4.02 \times 10^{-2}$ & $2.44 \times 10^{-2}$ & $2.67 \times 10^{-2}$ & $1.28 \times 10^{-2}$ 
                      \\&&& St.d. & $6.01 \times 10^{-2}$ & $5.03 \times 10^{-2}$ & $3.09 \times 10^{-2}$ & $3.37 \times 10^{-2}$ & $1.57 \times 10^{-2}$ \\
                      \cline{3-9}
                    && \multirow{2}{*}{0.1} & MAE  & $1.03 \times 10^{-2}$ & $8.34 \times 10^{-3}$ & $5.73 \times 10^{-3}$ & $5.41 \times 10^{-3}$ & $2.78 \times 10^{-3}$ 
                      \\&&& St.d. & $1.23 \times 10^{-2}$ & $1.05 \times 10^{-2}$ & $7.13 \times 10^{-3}$ & $6.77 \times 10^{-3}$ & $3.48 \times 10^{-3}$ \\
                           \hline
                           
\multirow{6}{*}{\textbf{---}} & \multirow{6}{*}{Quintic Spline} & \multirow{2}{*}{1.0}                             & MAE  & $6.59 \times 10^{-2}$ & $4.57 \times 10^{-2}$ & $3.77 \times 10^{-2}$ & $3.98 \times 10^{-2}$ & $2.12 \times 10^{-2}$ \\
                      &&& St.d. & $8.04 \times 10^{-2}$ & $5.75 \times 10^{-2}$ & $4.69 \times 10^{-2}$ & $4.95 \times 10^{-2}$ & $2.64 \times 10^{-2}$ \\
                      \cline{3-9}
                      && \multirow{2}{*}{0.5} & MAE  & $3.44 \times 10^{-2}$ & $2.42 \times 10^{-2}$ & $1.98 \times 10^{-2}$ & $2.09 \times 10^{-2}$ & $1.10 \times 10^{-2}$ 
                      \\&&& St.d. & $4.24 \times 10^{-2}$ & $3.03 \times 10^{-2}$ & $2.46 \times 10^{-2}$ & $2.59 \times 10^{-2}$ & $1.38 \times 10^{-2}$ \\
                      \cline{3-9}
                    && \multirow{2}{*}{0.1} & MAE  & $7.01 \times 10^{-3}$ & $4.98 \times 10^{-3}$ & $4.05 \times 10^{-3}$ & $4.26 \times 10^{-3}$ & $2.23 \times 10^{-3}$ 
                      \\&&& St.d. & $8.63 \times 10^{-3}$ & $6.19 \times 10^{-3}$ & $5.04 \times 10^{-3}$ & $5.27 \times 10^{-3}$ & $2.80 \times 10^{-3}$ \\
                           \hline

\end{tabular}
\end{center}
\end{table}
The cross-consistency analysis in Table~\ref{tab:noisy_moments} demonstrates a clear trade-off between peak specialization and operational robustness. As expected, the operator trained on minimal disorder ($\epsilon= 0.1$) achieves the lowest total MAE for the low-order moments ($2.04 \times 10^{-4}$) when evaluated in its native noise. However, this model is highly specialized; when subjected to higher disorders ($\epsilon \in \{0.5,1.0\}$), this model encounters out-of-distribution geometric configurations that were not represented during training, leading up to a two-order-of-magnitude increase in residuals.

In contrast, training on the highest noise level ($\epsilon = 1.0$) provides a broader coverage of stencil arrangements, including the near-regular states found at lower noise levels, the resulting model functions as a robust generalist. While this breadth prevents the operator from reaching the same high precision on a more structured point cloud, it ensures that the moments remain well-satisfied for a wide range of particle distribution. For Lagrangian simulations, where particle distributions inevitably transition from ordered to highly disordered states, the distributional robustness afforded by high-noise training is a critical prerequisite for maintaining numerical consistency throughout the simulation.

\begin{figure}[ht]
  \begin{center}
  \includegraphics[width=0.8\linewidth]{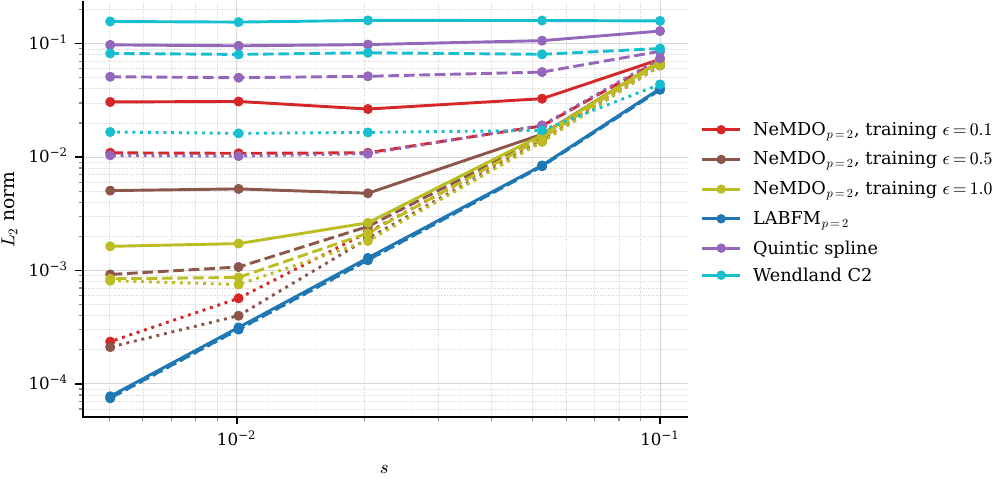}
  \caption{Convergence of the discrete $x$-derivative operator on a smooth test function (Equation~\eqref{eq:test_func}) for LABFM$_{p=2}$, SPH (with the quintic spline and Wendland C2 kernels) and NeMDO,  with varying particle disorder $\epsilon$. Full lines indicate inference with $\epsilon=1.0$, dashed lines indicate inference with $\epsilon=0.5$, and dotted lines indicate inference with $\epsilon=0.1$.}
    \label{fig:noisy_conv}
  \end{center}
\end{figure}

Figure~\ref{fig:noisy_conv} illustrates the $L_2$ error norm for the $x$-derivative operator across varying spatial resolutions and degrees of particle disorder. Across all noise regimes, both SPH kernels exhibit a significantly higher truncation error floor compared to the NeMDO and LABFM frameworks. For the majority of operators, the asymptotic error limit is strongly correlated with the magnitude of particle disorder, with reduced stochasticity leading to superior accuracy.

An exception is observed in the LABFM results, which maintain a relatively invariant profile regardless of the disorder level. Given that LABFM explicitly enforces consistency by solving local linear systems at each evaluation point, it remains numerically robust even in highly disordered configurations where a valid solution to the moment equations exists. This behavior aligns with the extensive analysis of particle disorder and approximation order conducted by \citet{King2020}.

The NeMDO results further substantiate the distributional coverage observations from Table~\ref{tab:noisy_moments}. Operators trained under high perturbations ($\epsilon = 1.0$) demonstrate a convergence plateau that remains largely independent of the inference noise level, provided the latter does not exceed the training threshold. Conversely, models trained on low-disorder distributions exhibit a significant increase in the limiting error when applied to particle configurations that lie outside their training manifold. While LABFM achieves robustness by solving a linear system for each local neighborhood, NeMDO attains a similar robustness by training on a wide ensemble of disordered configurations ($\epsilon = 1.0$).

\section{Limitations} 
\label{sec:limitations}
In its present form, each model is trained for a fixed stencil size $|\mathcal{N}_i|$ and does not natively support variable-size neighbourhoods at inference time. Currently, this lack of topological flexibility necessitates the training of discrete operators for every desired neighbor count. Allowing variable-size neighbourhoods would provide greater flexibility and simplify integration with varying particle distributions. 

Another limitation of the current framework is that the achievable accuracy degrades as particle disorder increases. This behaviour reflects the growing difficulty of satisfying polynomial moment constraints under highly irregular stencils and is not unique to NeMDO: classical SPH discretisations and consistency-corrected mesh-free methods similarly suffer from reduced accuracy and conditioning issues under strong node disorder at higher reproduction orders. These effects become especially relevant in Lagrangian settings, where particle densities and neighbourhood sizes can vary significantly over time. In such cases, the framework would likely need to be coupled with particle-shifting or regularisation strategies~\citep{Xu2009ISPH,Lind2012ISPH, Oger2016}, as is standard practice in SPH-based methods, to maintain suitable stencil quality. Extending the architecture to more robustly handle adaptive stencils and highly disordered configurations is an important direction for future work.


\section{Conclusion \& Future Work}
\label{sec:conclusion}
This work presented NeMDO, a self-supervised learning framework for the construction of mesh-free discrete differential operators on unstructured particle sets. By training graph neural networks to satisfy polynomial moment constraints derived from truncated Taylor expansions, we have developed a methodology for generating operators that are physics-agnostic, strictly local, and robust to significant geometric disorder. Through multiple benchmarks—including moment and stability analysis, convergence studies, modal response, and Taylor–Green vortex simulations---we demonstrated that NeMDO operators achieve superior accuracy compared to standard SPH discretizations. Furthermore, the learned operators remain competitive with a representative consistent method across a range of resolutions, while offering a tunable accuracy-to-computational-cost trade-off that is not readily accessible to traditional mesh-free formulations.

While the current study focuses primarily on the fundamental properties of the discrete operators and their weight-generation consistency, rather than the implementation of complex boundary conditions or adaptive refinement strategies, it provides the necessary mathematical foundation for more sophisticated mesh-free applications. More broadly, this research suggests that learning spatial discrete operators provides a principled pathway for integrating machine learning into hybrid mesh-free PDE solvers.

The present implementation utilizes a message-passing GNN architecture, however, the NeMDO framework is compatible with multiple permutation-invariant architectures, including Transformers~\citep{lin2021surveytransformers}, DeepSets~\citep{zaheer2018deepsets}, and Equivariant Neural Networks~\citep{Batzner2022}. Consequently, future research will benchmark alternative architectures to optimize the balance between expressive power and inference speed. Other directions for future work include: (i) extending the NeMDO framework to complex geometries and Lagrangian flows, where the robustness of the learned operators to extreme topological deformation can be fully leveraged, (ii) investigate training objectives that incorporate spectral or modal targets directly into the loss function to further refine high-wavenumber resolving power, and (iii) integrate symbolic regression for discovering compact kernel functions that retain the robustness of learned operators while offering the interpretability of classical mesh-free methods.

\section*{Data and Code Availability}
The complete implementation for solver development and neural network training, along with the datasets used for training and the scripts for convergence, stability, and resolving power analyses, is available as open-source at \url{https://github.com/uom-complexfluids/nemdo}. The repository is organized into modular directories for training and testing, including all benchmarked methods and hyperparameter configurations used in this study.

\section*{Declaration of Interest}
The authors declare that they have no known competing financial interests or personal relationships that could have appeared to influence the work reported in this paper.

\section*{Acknowledgements}
Jack R. C. King is funded by the Royal Society via a University Research Fellowship (URF\textbackslash R1\textbackslash 221290). We would like to acknowledge Dr. Henry Broadley for his insightful discussions, and for assistance given by Research IT and the use of the Computational Shared Facility at the University of Manchester.

\renewcommand{\refname}{Bibliography}
\bibliographystyle{elsarticle-num-names} 
\bibliography{bibliography}

\appendix

\section{Model Hyper-parameters and Training Dataset}
\label{app:hyper_p}
Below we summarise the NeMDO architectures used in the different experimental sections.

\paragraph{Main experiments} For the polynomial consistency and convergence (Section~\ref{sec:convergence}), modal response (Section~\ref{sec:modal_resp}), Taylor--Green vortex simulations (Section~\ref{sec:tgv}), and stability (Section~\ref{sec:stability}), we use a single architecture for both the $x$-derivative and Laplacian operators, denoted NeMDO$_{p=2}^x$ and NeMDO$_{p=2}^\Delta$. Their hyper-parameters are listed in Table~\ref{tab:hyp1}. In the computational cost-accuracy plot (Figure~\ref{fig:error_time}), this configuration is referred to as NeMDO$_{p=2}^3$. The models presented in table~\ref{tab:hyp1} were trained with approximately $7$ million neighbourhood graphs, with about $2$ million samples used for validation and $1$ million reserved for testing, yielding a total of roughly $10$ million neighbourhoods. The models were trained with a starting learning rate of $1 \times 10^{-5}$ and a plateau scheduler for a total of 2,000 epochs.

\paragraph{Parametric study.} During the parametric study on Taylor truncation and stencil size (Section~\ref{sec:ablation}), we fix the network architecture and vary only the reproduction order $p \in \{2,3\}$ and the number of neighbours $|\mathcal{N}_i| \in \{10,25,50,100\}$. All models in this group share the same parameter count and embedding dimension. The models have 183.7k trainable parameters, all MLPs (encoder, decoder, message and update) only have 1 hidden layer, the GNN has 2 graph layers, and all models with an embedding of size 128, with particle disorder of $\epsilon=1.0$, trained for 1,000 epochs, with a starting learning rate of $1 \times 10^{-4}$ with a plateau scheduler. The models presented in the parametric investigation were trained with approximately 115k neighbourhood graphs, with about 33k samples used for validation and 16k reserved for testing, yielding a total of roughly 164k neighbourhoods. 

\paragraph{Cost–accuracy experiments.} For the computational cost and accuracy study in Figure~\ref{fig:error_time}, we use smaller-capacity models to probe the trade-off between parameter count, runtime, and limiting error. The architectures are summarised in Table~\ref{tab:hyp3}, all models were trained for 2,000 epochs. NeMDO$_{p=2}^3$ (Figure~\ref{fig:error_time}) corresponds to NeMDO$_{p=2}^x$ listed in Table~\ref{tab:hyp1}. The models presented in the computational cost-accuracy experiments (Table~\ref{tab:hyp3}) were trained with approximately 115k neighbourhood graphs, with about 33k samples used for validation and 16k reserved for testing, yielding a total of roughly 164k neighbourhoods.

\begin{table}[H]
\caption{NeMDO architectures used in the main experiments.}
\label{tab:hyp1}
\begin{center}
\begin{small}
\begin{tabular}{ccccccc}
\multicolumn{1}{c}{\bf Operator} &
\multicolumn{1}{c}{\bf $|\mathcal{N}_i|$} &
\multicolumn{1}{c}{\bf \# Parameters} &
\multicolumn{1}{c}{\bf $F_h$} &
\multicolumn{1}{c}{\bf \shortstack{MLP Hidden Layers \\(Emb/Out/Msg/Upd)}} &
\multicolumn{1}{c}{\bf \shortstack{Graph\\Layers}} &
\multicolumn{1}{c}{\bf $\epsilon$}
\\ \hline \\
NeMDO$_{p=2}^x$      & 35  & 2.2M   & 256 & 3 & 3 & 1.0 \\
NeMDO$_{p=2}^\Delta$ & 35  & 2.2M   & 256 & 3 & 3 & 1.0 \\
\end{tabular}
\end{small}
\end{center}
\end{table}

\begin{table}[H]
\caption{NeMDO architectures used in the cost–accuracy experiments.}
\label{tab:hyp3}
\begin{center}
\begin{small}
\begin{tabular}{ccccccc}
\multicolumn{1}{c}{\bf Operator} &
\multicolumn{1}{c}{\bf $|\mathcal{N}_i|$} &
\multicolumn{1}{c}{\bf \# Parameters} &
\multicolumn{1}{c}{\bf $F_h$} &
\multicolumn{1}{c}{\bf \shortstack{MLP Hidden Layers\\(Emb/Out/Msg/Upd)}} &
\multicolumn{1}{c}{\bf \shortstack{Graph\\Layers}} &
\multicolumn{1}{c}{\bf $\epsilon$}
\\ \hline \\
NeMDO$_{p=2}^{1}$   & 10  & 11.1k  & 32  & 1 & 2 & 1.0 \\
NeMDO$_{p=2}^{2}$   & 15  & 46.3k  & 64  & 1 & 2 & 1.0 \\
\end{tabular}
\end{small}
\end{center}
\end{table}

\section{Computational Cost Details}
\label{app:comp_cost}
Providing a fully balanced performance comparison between fundamentally different approaches is inherently difficult. The estimates in Figure~\ref{fig:error_time} should therefore be interpreted as indicative rather than absolute. All timings were obtained on a single Intel Core i9-14900F core (single thread). For the SPH baselines, we measure the time required to compute weights with the kernels. For LABFM, we include the time to assemble and solve the resulting linear systems, and form the final stencil weights. For NeMDO, we measure the wall-clock time of inference---we do not include the training time since the learned operator can be trained offline once, and implemented on different PDEs. The time measurements provided in Figure~\ref{fig:error_time} represents the total time for each operator to predict the $x$-derivative weights for all nodes in a given resolution.

Most neural-network components in our framework are implemented using deep learning libraries optimised for GPU and batched execution, whereas small dense linear systems can be solved very efficiently on CPUs. As a result, the single-core CPU timings reported here are conservative for NeMDO and relatively favourable for LABFM. We do not report GPU or multi-threaded timings in this work; a systematic study of heterogeneous CPU–GPU execution and distributed inference is left for future work. The key takeaway from the present measurements is the relative ordering and scaling behaviour of the methods: NeMDO eliminates the per-stencil linear solves required by corrected-kernel schemes and exposes explicit architectural and stencil-size knobs to trade compute for accuracy within a single framework.

\end{document}